\documentclass[journal]{IEEEtran}

\usepackage{graphicx} 
\usepackage{amsthm}
\usepackage{amssymb}
\usepackage{amsmath}
\usepackage{dblfloatfix} 

\pdfoutput=1

\ifCLASSINFOpdf
\else
\fi
\begin{document}
%
\title{Tunably Rugged Landscapes with Known Maximum and Minimum}
%
%
%

\author{Narine~Manukyan, %
        Margaret~J.~Eppstein, %
        and~Jeffrey~S.~Buzas 
\thanks{Narine Manukyan and Margaret J. Eppstein are with the Department
of Computer Science, University of Vermont, Burlington,
VT, 05401 USA e-mail: Narine.Manukyan@uvm.edu and Maggie.Eppstein@uvm.edu.}
\thanks{Jeffrey S. Buzas is with the Department and Mathematics and Statistics, University of Vermont, Burlington,
VT, 05401 USA e-mail: Jeff.Buzas@uvm.edu.}
}

\maketitle

\begin{abstract}

We propose {\em NM} landscapes as a new class of tunably rugged benchmark problems.  {\em NM} landscapes are well-defined on alphabets of any arity, including both discrete and real-valued alphabets, include epistasis in a natural and transparent manner, are proven to have known value and location of the global maximum and, with some additional constraints, are proven to also have a known global minimum. Empirical studies are used to illustrate that, when coefficients are selected from a recommended distribution, the ruggedness of {\em NM} landscapes is smoothly tunable and correlates with several measures of search difficulty. We discuss why these properties make {\em NM} landscapes preferable to both {\em NK} landscapes and Walsh polynomials as benchmark landscape models with tunable epistasis.

\end{abstract}


%
\IEEEpeerreviewmaketitle

\section{Introduction}
%
%
%
%
\IEEEPARstart{S}{imulated } 
landscapes are widely used for evaluating search strategies, where the goal is to find the landscape location with maximum fitness value \cite{kauffman1989nk}\cite{eppstein2012searching}. Without loss of generality and for notational simplicity, we assume function maximization, rather than minimization, throughout this paper.

 {\em NK} Landscapes \cite{kauffman1989nk} have been classic benchmarks for generating landscapes with epistatic interactions. They are described by two parameters: {\em N} specifies the number of binary features and $K$ specifies that the maximum degree of epistatic interactions among the features is $K+1$ \cite{stuart1993origins}. {\em NK} landscapes have been used in many applications (e.g., \cite{giannoccaro2013complex,moraglio2009geometric,rowe2010analysis,tomko2013unconstrain,arnaldo2013matching}) and widely studied in theory (e.g., \cite{pelikan2008analysis,wright2000computational,buzas2013analysis,hordijk2014correlation,liaw2013effect}), as they can generate landscapes with tunable ruggedness by varying $K$. 
However, {\em NK} landscapes have several limitations. Buzas and Dinitz \cite{buzas2013analysis} recently showed that the expected number of local peaks in {\em NK} landscapes rises in large discrete jumps as $K$ is increased, but actually decreases as a function of the number of interaction terms for a given $K$ (Fig. 1, red lines).  Additionally, the problem of finding the location and value of the global optimum of unrestricted {\em NK} landscapes with $K > 1$ is NP-complete \cite{wright2000computational} (although for restricted classes one can use  dynamic programming \cite{wright2000computational}\cite{gao2002analysis}  or approximation algorithms \cite{wright2000computational}). 
{\em NK} landscapes have only been defined for binary alphabets.

Walsh polynomials are a superset of {\em NK} landscapes that overcome some of the limitations of {\em NK} landscapes. For example, they allow more explicit control over which interaction terms are present. The problem of finding the global maximum value of a Walsh polynomial is also NP-complete, although a restricted subset of Walsh polynomials has a known global maximum \cite{tanese1989distributed}. However, even in this case finding the global minimum is still NP-complete, preventing proper normalization by the range of fitnesses in the landscapes. 
As with {\em NK} landscapes, Walsh polynomials are only defined for binary alphabets. 

\begin{figure}[h!]
\includegraphics[width=3.5in]{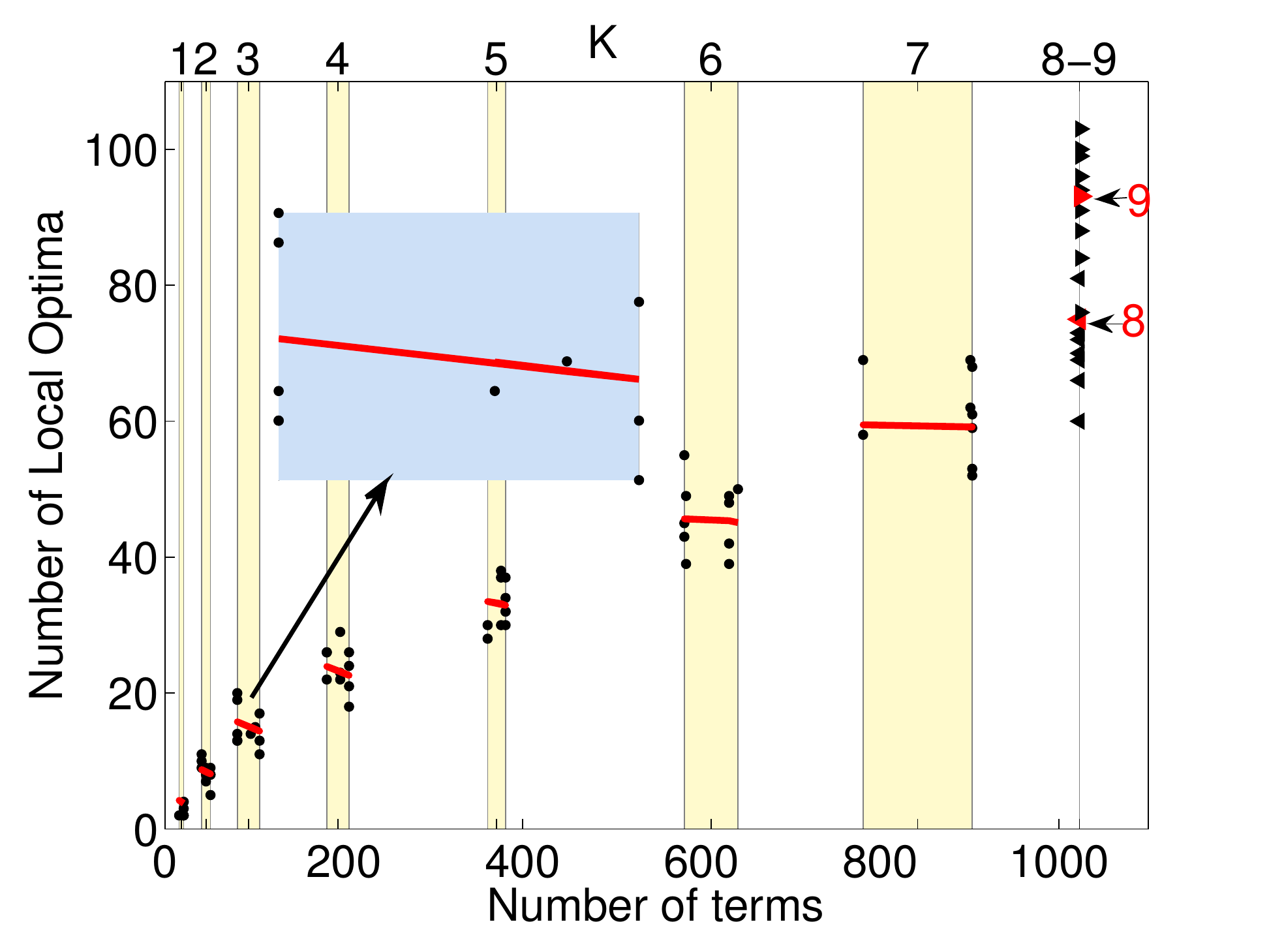}
\vspace{-20pt}
\caption{Number of local peaks in {\em NK} landscapes with $N=10$, as a function of the number of terms in the equivalent parametric interaction model ($m$, bottom x-axis) for $K =\{1,2,...,9\}$ (top x-axis). The black dots show empirical results of 10 random landscapes generated for each value of $K$; red lines show the expected number of peaks ($L$) of these same landscapes computed according to the formula given in \cite{buzas2013analysis}. The inset shows a magnification of the $K = 3$ results.}
\label{fig1}
\vspace{-5pt}
\end{figure}

In this paper, we introduce a different, more flexible subset of general interaction models that we dub {\em NM} landscapes. Like {\em NK} landscapes and Walsh polynomials, {\em NM} landscapes incorporate epistatic feature interactions. However, {\em NM} landscapes also (a) include epistasis in a natural and transparent manner, (b) have known value and location of the maximum fitness, (c) work with alphabets of any arity, including discrete and real-valued alphabets, (d) with additional constraints have known value and location of the minimum fitness, and (e) when coefficients are chosen properly, have relatively smoothly tunable ruggedness. In Section 2 we introduce the general class of parametric interaction models and Walsh polynomials, then in Section 3 we define {\em NM} landscapes and prove the properties (a), (b), (c) and (d) above. In Sections 4 and 5 we describe experiments and results that demonstrate property (e) above.  In Section 6 we discuss the importance of these properties and point out several advantages of {\em NM} landscapes as benchmark problems for studying search in tunably rugged landscapes.

\section{Interaction Models and Walsh Polynomials}
Walsh polynomials provide a mathematical framework for defining any real-valued function on bit strings \cite{mitchel91}\cite{kallel2001properties}. A Walsh polynomial has the following form:
\begin{equation}
f({\bf y}) = \sum_{j=0}^{2^q-1}\omega_j\psi_j({\bf y})
\label{Walsh1}
\end{equation}
where $q$ is the length of the bit string ${\bf y}$,  each bit $y_i\in\{0,1\}$, and each $\omega_j \in \mathbb{R}$. The Walsh function $\psi_j({\bf y})$  corresponding to the $j$th partition is defined as:
\begin{equation}
\psi_j({\bf y}) = \begin{cases} 1, & \mbox{if } {\bf y} \wedge j_2 \mbox{ has even parity}\\
-1, & \mbox{otherwise} \end{cases}
\label{Walsh2}
\end{equation}
where $j_2$ denotes the binary representation of $j$.

{\em NK} landscapes are a subset of Walsh polynomials. Walsh polynomials have a one-to-one correspondence with the more general class of general parametric interaction models, when such models are restricted to binary alphabets \cite{kallel2001properties}. 

 A fitness landscape $F$ can be defined for $N$ features using a general parametric interaction model of the form:
\begin{equation}
F({\bf x}) = \sum_{k=1}^m \beta_{U_k}\prod_{i\in U_k}x_i
\label{Interaction1}
\end{equation}
where $m$ is the number of terms, and each of $m$ coefficients $\beta_{U_k}  \in \mathbb{R}$. For $k = 1\dots m$, $U_k \subseteq \{1,2,\dots,N\}$, where $U_k$ is a set of indices of the features in the $k$th term, and the length $|U_k|$ is the order of the interaction. We adopt the convention that when $U_k = \emptyset$, $\prod_{j \in U_k}x_j \equiv 1$. If the parametric interaction model is defined on a binary alphabet, we adopt the convention that binary values are represented as $x_i \in \{-1,1\}$ (rather than $\{0,1\}$, as in Walsh polynomials).  However general parametric interaction models are also well defined for discrete valued features with higher arities as well as for real-valued alphabets and provide a more intuitive way of representing epistatic interactions among features.
 \vspace{2pt}

A more readable notation for Eq. (\ref{Interaction1}) is as follows:
\begin{align} 
F({\bf x}) &=  \beta_0 +  \sum_{i = 1}^{N}\beta_i x_i + \nonumber \\
& + \sum_{i = 1}^{N-1} \sum_{j = i+1}^N \beta_{i,j}x_i x_j+ \nonumber \\
&  + \sum_{i = 1}^{N-2} \sum_{j = i+1}^{N-1} \sum_{k = j+1}^N\beta_{i,j,k}x_i x_j x_k+H
\label{Interaction2}
\end{align}
where we only explicitly show up to third order interactions and $H$ represents  higher order interactions up to some maximum order $M \le N$.
Note that some $\beta_{U_k}$ parameters may be zero, so not all terms need be present.

For example, consider a model with $N = 2$ loci and $U_1=\emptyset$, $U_2 = \{1\}$, $U_3 = \{2\}$ and $U_4 = \{1,2\}$. The interaction model for this example is:
\begin{equation}
F({\bf x}) = \beta_0+\beta_1x_1+\beta_2x_2+\beta_{1,2}x_{1}x_2
\label{interactionExample}
\end{equation}
where $\beta_0$ is the average value of all fitnesses in the landscape, $\beta_1$ and $\beta_2$ are the coefficients of the main effects of the binary features $x_1$ and $x_2$, and $\beta_{1,2}$ is the coefficient of the second order epistatic interaction $x_{1}x_2$. 
The Walsh polynomial corresponding to Eq. (\ref{interactionExample}) is:
\begin{align}
 \notag f({\bf y}) = \omega_0\psi_0({\bf y}) + \omega_1\psi_1({\bf y}) + \omega_2\psi_2({\bf y}) + \omega_3\psi_3({\bf y}) \\
=  \beta_0\psi_0({\bf y}) - \beta_1\psi_1({\bf y}) - \beta_2\psi_2({\bf y}) + \beta_{1,3}\psi_3({\bf y})
\label{equivalence}
\end{align}
where
\begin{equation}
y_i = \begin{cases} 1, &\mbox{  when } x_i = 1\\ 0, & \mbox{  when } x_i = -1
\end{cases}
\end{equation}
Notice that there is a one-to-one correspondence of each term in Eq.  (\ref{interactionExample}) with each term in Eq. (\ref{equivalence}) but the signs of the coefficients are different. Specifically, for the example above:
\begin{equation}
\beta_0 = \omega_0,\hspace{4pt}  \beta_1 = - \omega_1, \hspace{4pt} \beta_2 = - \omega_2, \hspace{4pt} \beta_{1,2} =  \omega_3
\label{betas}
\end{equation}

A random point selected in the search space of a Walsh polynomial can be forced to be the global maximum by properly adjusting the sign of each of the non-zero Walsh coefficients, with the maximum fitness value equal to the sum of the absolute values of all Walsh coefficients \cite{tanese1989distributed}.
However, the location and the value of the global minimum remains unknown.

General parametric interaction models are the standard models used in statistics to study effects of multiple features on an outcome (e.g., \cite{montgomery1984design}). They are easy to define and the interactions are transparent and easy to interpret (unlike in {\em NK} landscapes and Walsh polynomials). For example, the interaction terms present in Eq. (\ref{interactionExample}) are clearly evident, whereas the Walsh functions $\psi_i$ in Eq. (\ref{equivalence}) obscure this. To date, general parametric interaction models have received very little attention in the evolutionary computation literature, with notable exceptions \cite{reeves2000experiments,reeves1995epistasis,reeves1995experimental}.

In \cite{buzas2013analysis} the authors show that for every {\em NK} landscape with a given $K$ one can create an equivalent parametric interaction model, where the maximum order of interactions is $K+1$.  They show that the {\em NK} algorithm dictates that the interaction model contain all main effects and sub-interactions contained in higher order interactions. For example, if a non-zero interaction coefficient $\beta_{i,j,k}$ is present in an {\em NK} landscape, then there will generally be non-zero coefficients $\beta_i, \beta_j, \beta_k, \beta_{i,j}, \beta_{i,k}, \beta_{j,k}$ (there is an infinitesimally small probability that one or more of these coefficients may be zero). For the classic {\em NK} model where $K$ is constant and $K\ll N$, main effect coefficients have the largest expected magnitude, second order interactions have larger expected magnitude than third order interactions, and so on \cite{buzas2013analysis}. Thus, {\em NK} landscapes are a very restricted subset of  Walsh polynomials and the more general class of parametric interaction models.

\section{ {\em NM} Landscapes}
The class of Walsh polynomials is a subset of the larger class of general interaction models.  
Here we introduce a different subset of general interaction models called {\em NM} landscapes, where $N$ is the number of features and all interactions in the model are of order $\le M$. 

{\bf Definition 1:} {\em NM} models comprise the set of all general interaction models specified by Eq. (\ref{Interaction1}), with the added constraints that  (a) all coefficients $\beta_{U_k} $ are non-negative, (b) each feature value $x_i$ ranges from negative to positive values, and (c) the absolute value of the lower bound of the range $\le$ the upper bound of the range of $x_i$. 

\begin{figure}[h!]
\hspace{-10pt}\includegraphics[width=3.7in]{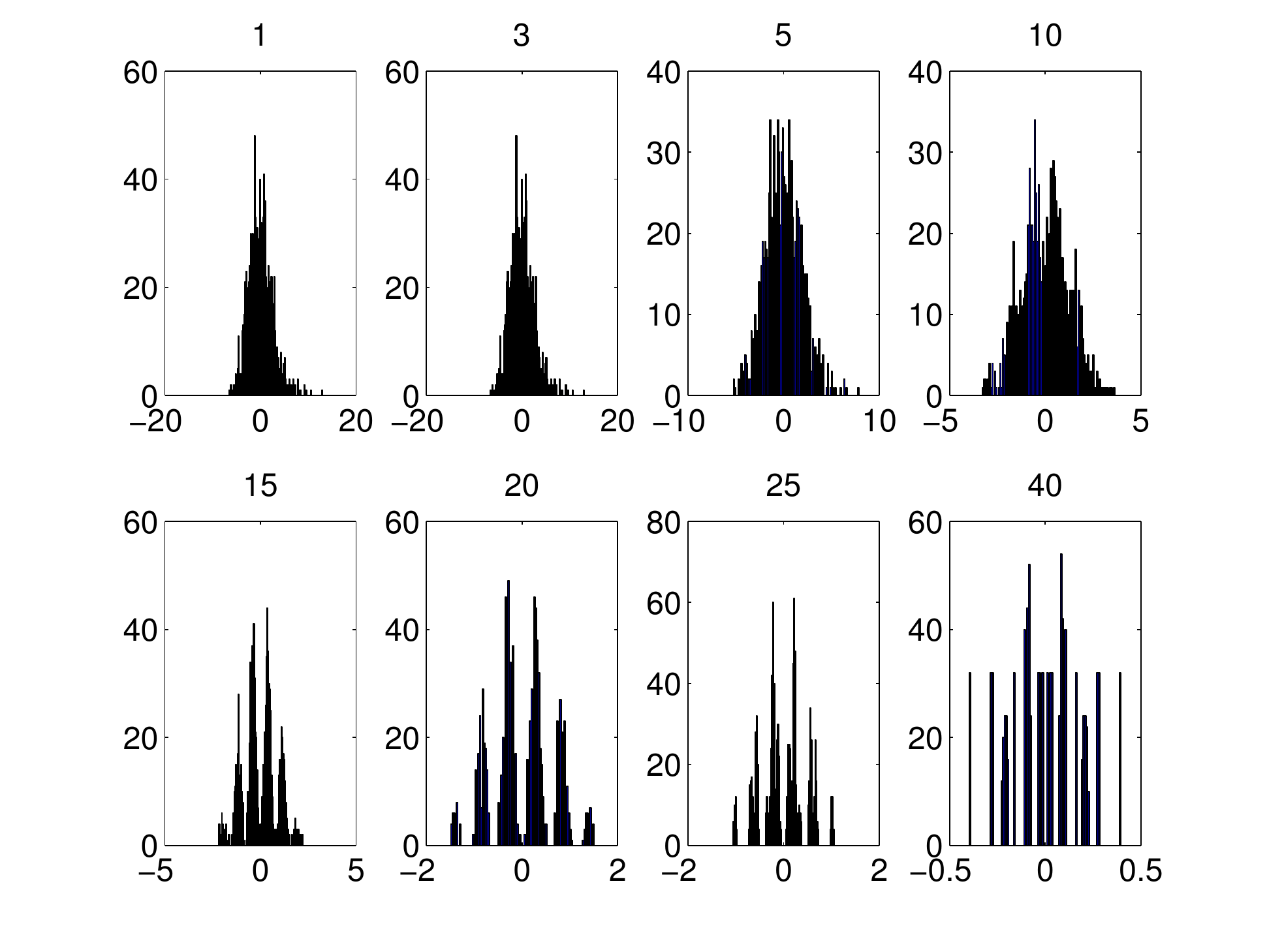}
\caption{Histograms of all 1024 fitnesses in {\em NM} landscapes for $M = 2$, $N = 10$ and coefficients drawn from Eq. (\ref{betadist}) with $\sigma$ as indicated.}
\label{histFigLabel}
\end{figure}

\begin{figure}[h!]
\hspace{50pt}\includegraphics[width=2in]{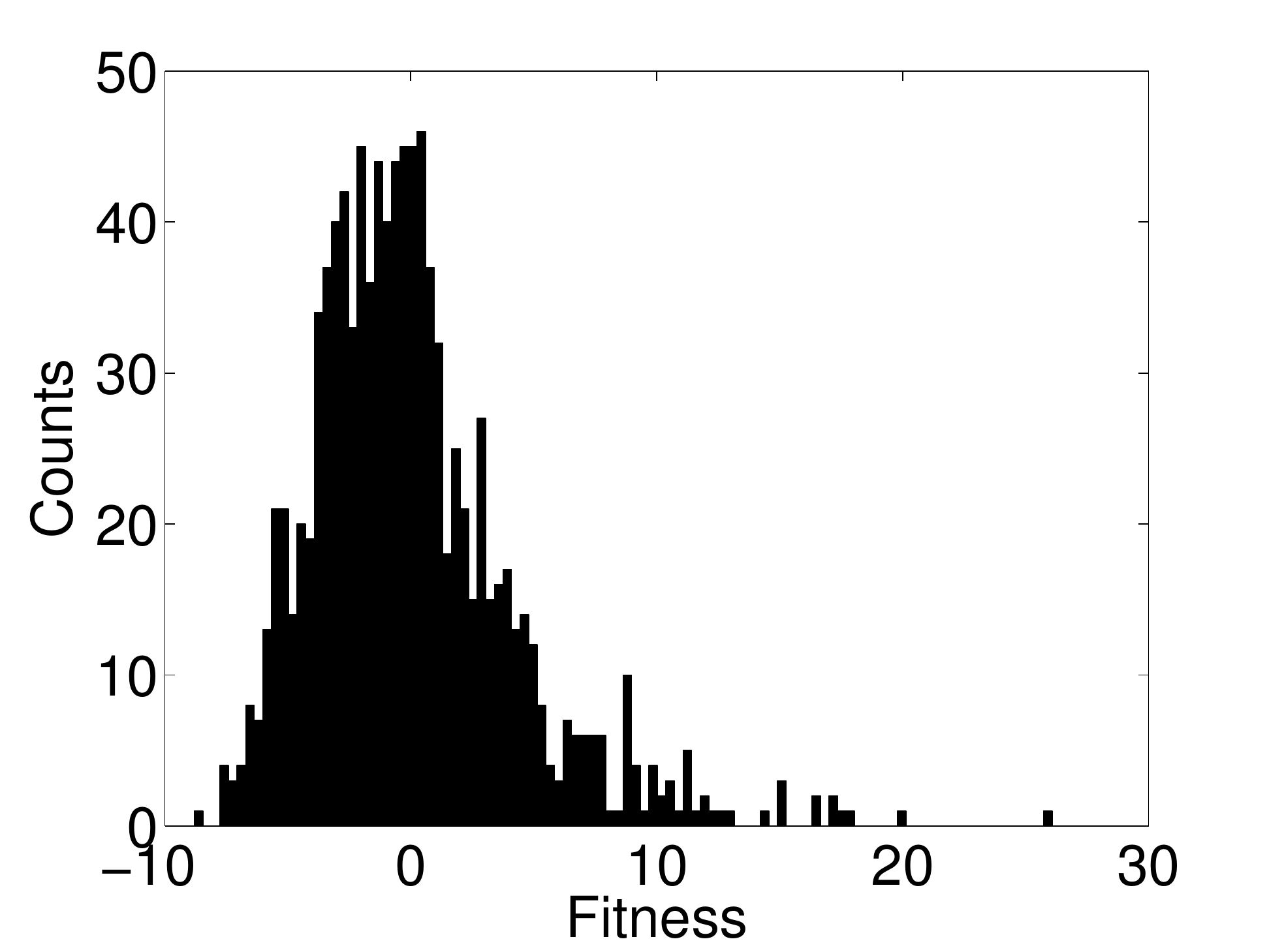}
\caption{Histograms of all 1024 fitnesses in binary {\em NM} landscapes for $M = 2$, $N = 10$ and coefficients drawn from a uniform distribution in the range $[0,1]$.}
\label{histFigLabel1}
\end{figure}

In this work, each $\beta_{U_k} $ is randomly created as follows:
 \begin{equation}
\beta_{U_k} = e^{-abs(\mathbb{N}(0,\sigma))}
\label{betadist}
\end{equation}
where $\mathbb{N}(0,\sigma)$ is a random number drawn from a Gaussian distribution with $0$ mean and standard deviation of $\sigma$, which results in fitnesses that are symmetric around $0$ (Fig. \ref{histFigLabel}). As the value of $\sigma$ increases, the means and standard deviations of the coefficients decrease, which results in a smaller range of fitness values and increasing clumping of fitness values (Fig. \ref{histFigLabel}). In contrast, when coefficients are drawn from a uniform distribution in the range $[0, 1]$, the fitnesses are skewed left (\ref{histFigLabel1}). {\em NM} landscapes offer several desirable properties, as described in the following.

{\bf Proposition 1:}
{\em NM} landscapes with a binary alphabet have a known global maximum.
\begin{proof}
By Definition 1, $\beta_{U_k}>0$ for all non-zero terms. Thus, the maximum possible value for each term ($\beta_{U_k} \prod_{j\in U_k}x_j$) in an {\em NM} landscape with a binary alphabet $x_i \in \{-1,1\}$ is achieved when:
\begin{equation}
x_i = 1, \hspace{10pt} \forall i = 1 \dots n
\end{equation}
and the value of the global maximum is:
\begin{align} 
F_{max} &=  \beta_0 +  \sum_{i = 1}^{N}\beta_i + \nonumber \\
& + \sum_{i = 1}^{N-1} \sum_{j = i+1}^N \beta_{i,j} + \nonumber \\
& +\sum_{i = 1}^{N-2} \sum_{j = i+1}^{N-1}  \sum_{k = j+1}^N \beta_{i,j,k} + \sum_{\forall \beta_{U_h}}\beta_{U_h}
\label{sumConsts}
\end{align}
where $\beta_{U_h}$ are the coefficients of all the remaining higher order interactions. Note that the calculation of Eq. (\ref{sumConsts}) has time complexity of $O(m)$, where $m$ is the number of terms in the model.
\end{proof}

{\bf Proposition 2:} {\em NM} landscapes can be defined on discrete alphabets of any arity or on real-valued alphabets, and the value and location of a global maximum is independent of the discretization of the alphabet.
\begin{proof}

By Definition 1, all coefficients are non-negative, therefore the maximum $F_{max}$ of an {\em NM} landscape with any discrete or real-valued alphabet $x_{a,b}$ defined in the range $[-a, b]$ where $0<a \le b$, occurs when $x_i=b, \forall i=1\dots N$ and its value is:
\begin{align}
 F_{max} &=  \beta_0 +  \sum_{i = 1}^{N}\beta_i b + \nonumber\\
& + \sum_{i = 1}^{(N-1)} \sum_{j = i+1}^N \beta_{i,j}b^2+\nonumber\\
& + \sum_{i = 1}^{(N-2)} \sum_{j = i+1}^{(N-1)} \sum_{k = k+1}^N \beta_{i,j}b^3 + \sum_{\forall \beta_{U_h}}\beta_{U_h}b^{|U_h|}
\label{maxval}
\end{align}
where $\beta_{U_h}$ are the coefficients of all the remaining higher order interactions and the lengths $|U_h|$ are the orders of these interactions. Thus the magnitude of $F_{max}$ is a function of $b$, but is independent of the arity of the alphabet. \end{proof}
 
For alphabets where $a = b$, $\beta_0$ represents the mean value of the landscape. We note that it is trivial to extend this proposition and proof to {\em NM} landscapes with heterogeneous alphabets (i.e., different ranges and/or arities for each feature variable), as long as the lower bound for each feature is negative, the upper bound is positive, and the absolute value of the lower bound is $\le$ the upper bound. However, for notational simplicity we only demonstrate the proof for homogeneous alphabets. We refer to the above described general {\em NM} landscapes as Type I {\em NM} landscapes.

We conjecture that changing the arity of features in {\em NM} landscapes does not change the number, locations, or values of the local peaks or global minima, because higher arity alphabets simply sample the same landscape at a higher resolution that interpolates between the local peaks. (Empirical data, not shown, supports this conjecture.) 

Note that interaction models with all non-negative interaction coefficients $\beta_{U_k}$, but no negative feature values,  generate unimodal landscapes. However, since alphabets in {\em NM} landscapes are defined to include both negative and positive features, {\em NM} landscapes have multiple local optima whenever any interaction terms are included.

\vspace{8pt}
{\bf Proposition 3:} 
 {\em NM} landscapes that include all main effects have exactly one global maximum.
\begin{proof}
By proposition 1, a maximum fitness $F_{max}$ of an {\em NM} landscape is achieved at point ${\bf x} = [b,b,\dots b]$. Let ${\bf y} =[y_1,y_2,\dots y_n]$, where {\bf y} $\neq$ {\bf x} (i.e, there exists at least one $i$ such that $y_i \neq b$). Since each $x_i \in [-a, b]$ and $a \le b$ (by Definition 1) then $\beta_i*y_i < \beta_i*b$ and 
the value of the interaction model at point {\bf y} will be strictly less than the value at point {\bf x}.
Thus, {\bf x} is the only global maximum. 
\end{proof}

{\bf Proposition 4:}
{\em NM} landscapes that include only even order terms with alphabets in the range [-b, b] are symmetric and have exactly two global maxima at maximal distance apart in feature space. 
\begin{proof}
Since $x_i^{2t} =(-x_i)^{2t},  \hspace{3pt}  \forall t \in \mathbb{I}$, then for all {\em NM} landscapes with only even order terms, $F({\bf x}) = F(-{\bf x})$ for each pair of points ${\bf x} = [x_1,x_2,\dots,x_N]$ and $-{\bf x} = [-x_1, -x_2, \dots, -x_N]$. Thus, {\em NM} landscapes with only even order interactions and alphabets in $[-b,b]$ range are symmetric and the two global maxima are at locations $[b,b,\dots b]$ and $[-b,-b,\dots -b]$, which are the maximum distance away from each other in the feature space.
\end{proof}

When the value of the global maximum of the landscape is known one can partially normalize fitnesses to the range $\le 1$ using the following formula:
\begin{equation}
F = \frac{F}{F_{max}}
\label{NORMBYMAX}
\end{equation}
However, proper normalization of fitnesses to the interval [0,1] also requires prior knowledge of the global minimum of the landscape, as follows:
\begin{equation}
F = \frac{F-F_{min}}{F_{max}-F_{min}}
\label{NORMBYMINMAX}
\end{equation}
To this end, we define subsets of {\em NM} landscapes that have a known global minimum. While there are many ways to do this, below we present two such subsets.
 
\begin{figure*}[h!]
\hspace{15pt}\includegraphics[width=3.3in]{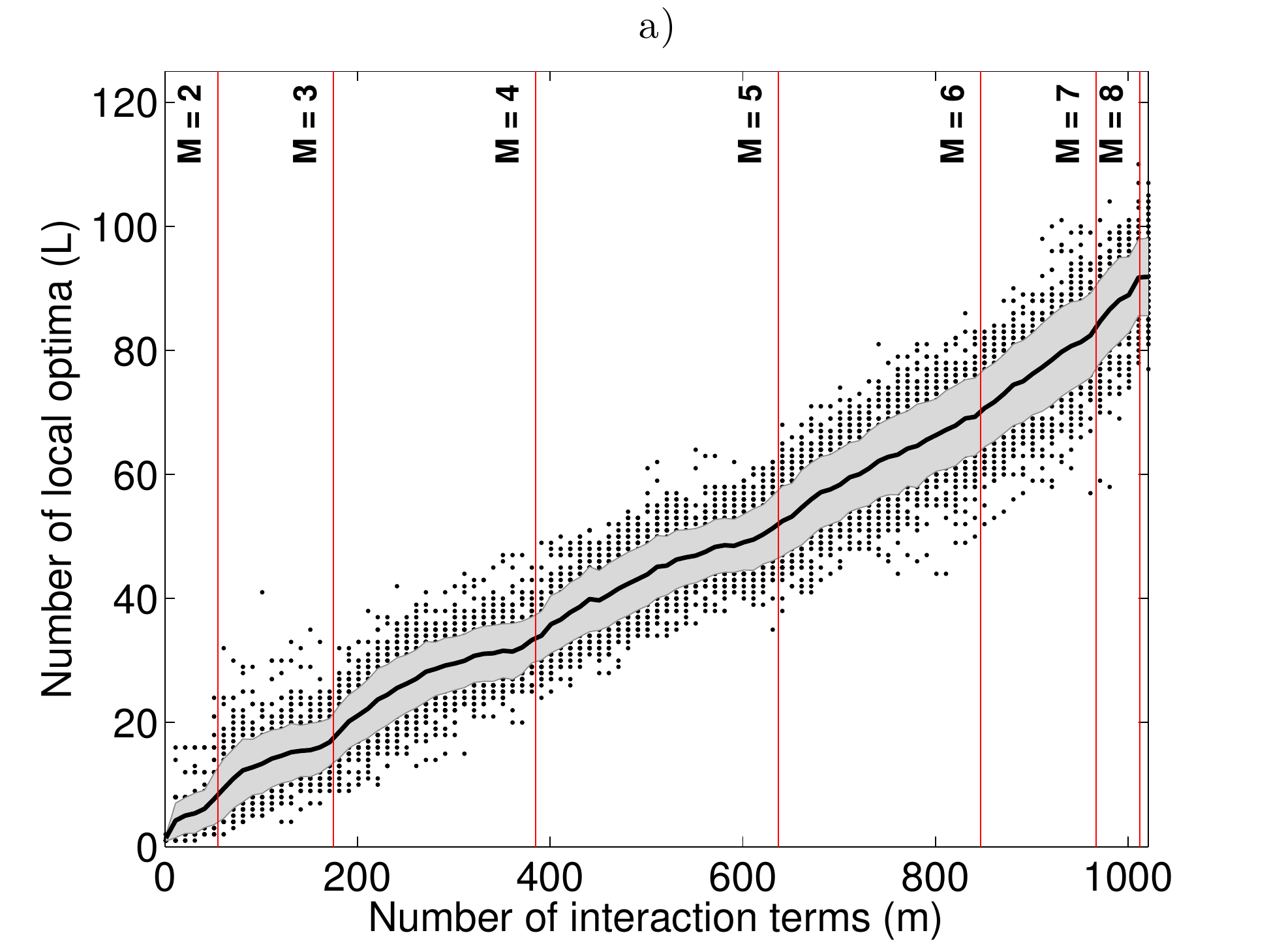}
\hspace{10pt}\includegraphics[width=3.3in]{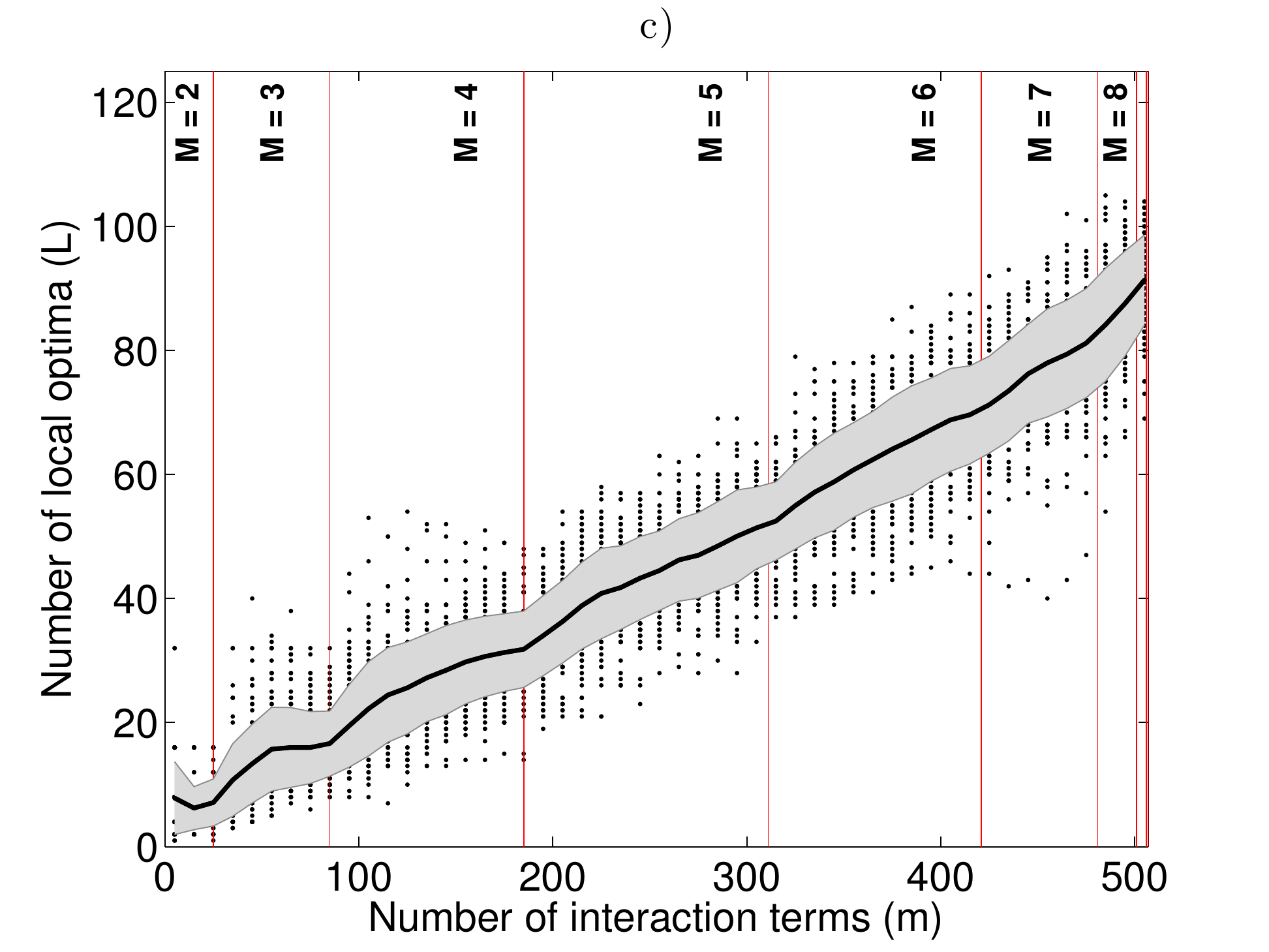}

\hspace{15pt}\includegraphics[width=3.3in]{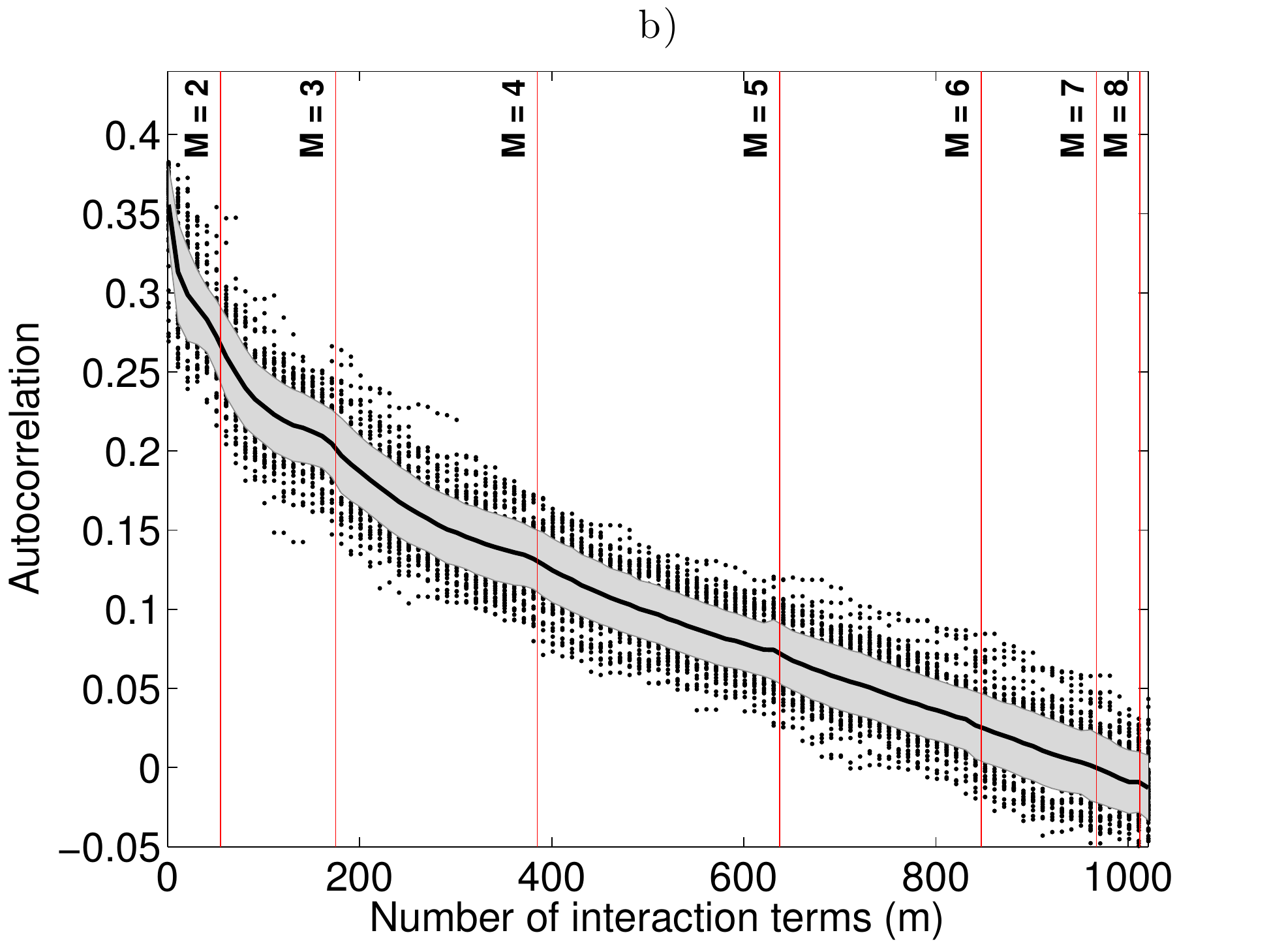}
\hspace{10pt}\includegraphics[width=3.3in]{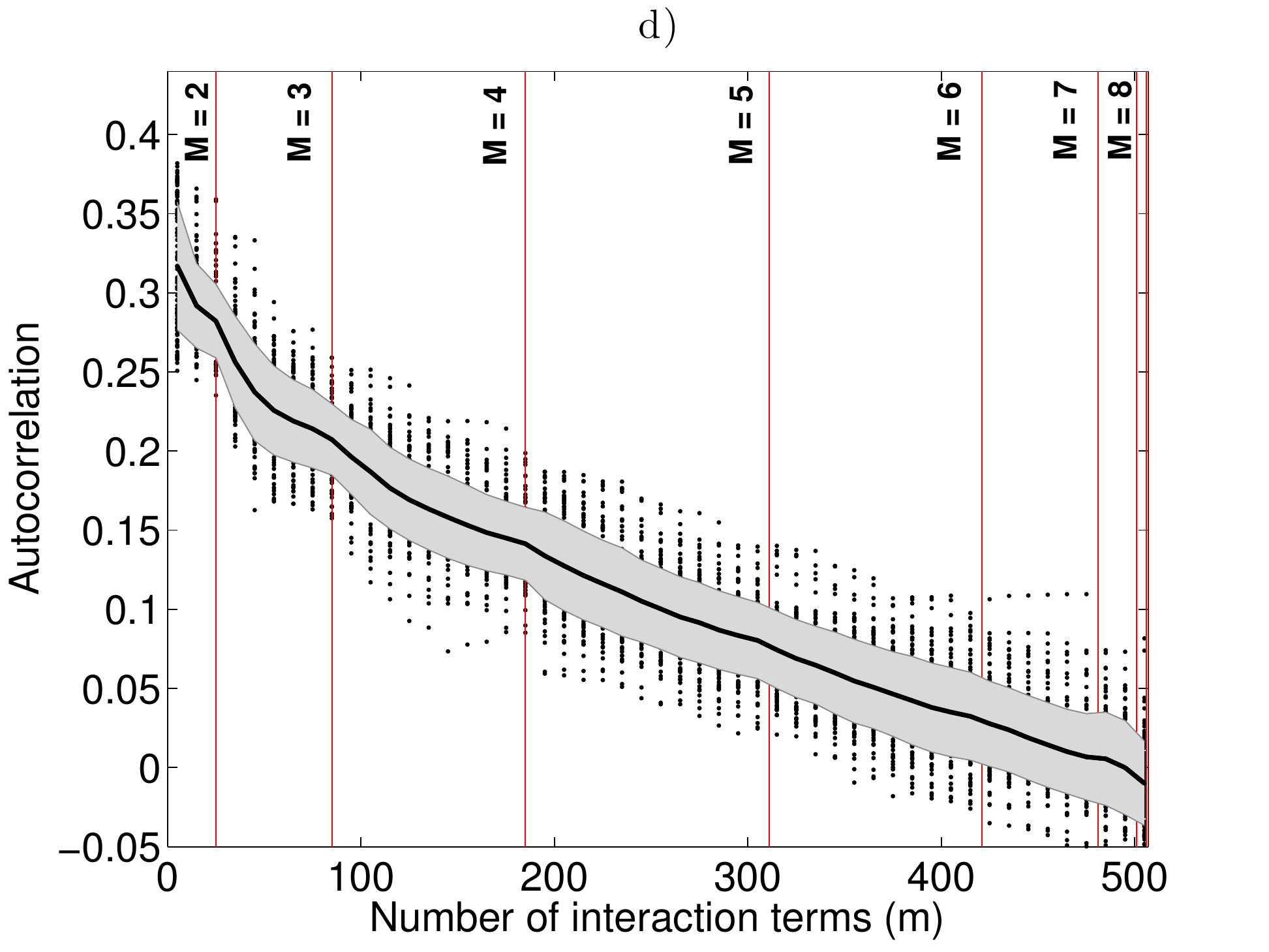}

\caption{Number of local peaks as a function of the number of terms $m$ (x-axis) and order of interactions $M$ (labels near top), for Type I (a)-(b) and Type II (c)-(d) binary {\em NM} landscapes with $N=10$ and $\sigma = 10$. The gray area shows one standard deviation and black lines show the means for 100 random {\em NM} landscapes. a) and c) show the number of local peaks, b) and d) show the lag 1 autocorrelation for the Type I and Type II {\em NM} landscapes respectively.}
\label{fig2}
\end{figure*}
 
{\bf Proposition 5:} 
 {\em NM} landscapes that include only main effects with odd indices (e.g., $x_1, x_3, x_5$, etc.) and any terms with an odd number of odd indices (e.g., $x_1x_2$, $x_1x_3x_5$, $x_1x_3x_6x_7$, etc.) and alphabets in the range $[-1,1]$ have a global minimum located at point $[-1,1,-1,1\dots]$. For example, models of this form including up to $M=3$ order terms are given by:

\begin{align}
\notag F(x) = &\beta_0 +  \sum_{i  \hspace{3pt} odd}^{N}\beta_ix_i +\sum_{\substack{i \hspace{3pt} odd,\\ j  \hspace{3pt} even}}^{N}\beta_{i,j}x_ix_{j}+\\
&\sum_{\substack{i  \hspace{3pt} odd,\\ j  \hspace{3pt} even,\\k  \hspace{3pt} even}}^{N}\beta_{i,j,k}x_ix_{j}x_k + \sum_{\substack{i  \hspace{3pt} odd,\\ j  \hspace{3pt} odd,\\k  \hspace{3pt} odd}}^{N}\beta_{i,j,k}x_ix_{j}x_k
\label{minModel}
\end{align}

 \begin{proof}

At the point $[-1,1,-1,1\dots]$ all terms with an odd number of odd indices will have a negative sign, as the product of an  odd number of negative numbers is negative. Thus, this point is the global minimum of the landscape with value:
\begin{equation}
F_{min} = -\big(\beta_0 +  \sum_{i  \hspace{3pt} odd}^{N}\beta_i + \sum_{\substack{i \hspace{3pt} odd,\\ j  \hspace{3pt} even}}^{N} \beta_{i,j} + \dots \big)
\end{equation}
(where only terms up through second order are explicitly shown above).
\end{proof}
We refer to the {\em NM} landscapes defined in Proposition 5 as Type II {\em NM} landscapes.  Note that Type II {\em NM} landscapes can easily be extended to alphabets in the range $[-a,b]$, where $a \le b$, although for notational simplicity we have limited the range to $[-1,1]$ in the above proof.
 
 \vspace{6pt}

{\bf Proposition 6:}
{\em NM} landscapes with only odd order terms and alphabets in the range $[-a,b]$, where $a \le b$, have a global minimum located at $[-a,-a,\dots,-a]$.

\begin{proof}
By Definition 1, $\beta_{U_k}>0$ for all non-zero terms, $x_i \in [-a,b]$ $\forall i$, and $a \le b$. Therefore the value of each term $T_k = \beta_{U_k}*x_{U_k}$ has to be $ \ge -|a|^{|U_k|}$. When all the features $x_i = -a$, $T_k = -|a|^{|U_k|}$. Therefore the point $[-a,-a,\dots,-a]$ is a global minimum with value:

\begin{align}
 F_{min} &=  -(\beta_0 +  \sum_{i = 1}^{N}\beta_i a + \nonumber\\
& + \sum_{i = 1}^{(N-2)} \sum_{j = i+1}^{(N-1)} \sum_{k = k+1}^N \beta_{i,j}a^3 + \sum_{\forall \beta_{U_h}}\beta_{U_h}a^{|U_h|})
\label{minval}
\end{align}

\end{proof}
We refer to the {\em NM} landscapes defined in Proposition 6 as Type III {\em NM} landscapes.
When $a = b$ the global maximum and minimum of Type III {\em NM} landscapes have the same absolute value, but opposite signs.
Because {\em NM} landscapes allow only non-negative coefficients but require both positive and negative feature values, we are thus able to construct {\em NM} landscapes with known maximum and known minimum, enabling normalization of fitnesses to the range [0,1] by equation (\ref{NORMBYMINMAX}).
In contrast, Walsh polynomials allow both positive and negative coefficients, but have only non-negative feature values. Thus, while it is possible to manipulate the signs of the Walsh coefficients to specify the location of the global maximum \cite{tanese1989distributed}, the global minimum of a Walsh polynomial is still unknown, even if one restricts the order of the interactions as in Type II or Type III {\em NM} landscapes.

\section{Experiments}
\subsection{Ruggedness}
We illustrate how ruggedness changes on binary {\em NM} landscapes with coefficients drawn from the distribution in Eq. (\ref{betadist}). Since we assess the ruggedness of these models using exhaustive search, we limit our experiments to $N \le 15$. 

In one set of experiments, we generated random Type I ``master" {\em NM} models, including terms for all $N$ main effects and the $\sum_{i=1}^{N}{N \choose i} $ possible interaction terms (e.g., for $N = 10$ there are 1023 overall terms; 1013 interaction terms plus 10 main effects). We then systematically created subsets of each of these master models that include an increasing number $m$ of terms from the master model, as follows.  We started with a base model that includes all main effects. Random second order terms were then added in groups of 10 (or less if there are not 10 left). After we had included all of the second order terms, we began adding randomly selected groups of 10 third order terms, and so on, until the single $N$-order interaction term was included.  We performed these incremental explorations of 100 master {\em NM} models for each of $N = 10$ with $\sigma = {10}$, and $N = 15$ with $\sigma \in \{15,20,100\}$.  

In another set of experiments we similarly created 100 master Type II {\em NM} landscapes according to Eq. (\ref{minModel}), with $N = 10$ and $\sigma = 10$. We created increasing subsets from the master models, as described above for the Type I landscapes.

We computed two standard measures of landscape ruggedness \cite{weinberger1990correlated,vassilev2000information}: (a) we counted the number of local peaks (where a local peak is defined as any point whose fitness value is greater than that of all of its neighbors); (b) we computed the lag 1 autocorrelation of random walks through the landscapes.

\subsection{Distribution of fitnesses and local peaks}

We generated representative {\em NM} landscapes with $N = 10$ and $\sigma = 10$ for each of $M \in \{1,2, 3, 4,6,10 \}$ for both Type I and Type II {\em NM} landscapes. We visualize these landscapes by plotting the fitnesses of all points in the landscape as a function of their distances (in feature space) to the global optimum, indicating which are local optima. 

\subsection{Basin of attraction of global optimum}
We assessed the size of the basin of attraction of the global maximum of Type I and Type III {\em NM} landscapes and {\em NK} landscapes for different values of $K = M-1 \in \{1,9\}$ and $N = 10$. The fitness matrix of {\em NK} landscapes is generated from random uniform numbers in the $[0, 1]$ range. We calculated the size of the global basin of attraction as a weighted sum of the points in the landscape that can reach the global maximum using only hill climbing.  Each point was weighted based on the percentage of its immediate neighbors with higher fitnesses that were also in the basin of attraction of the global maximum.

\subsection{Searchability of the landscapes}

We assessed how searchable {\em NM} landscapes are using simple genetic algorithms (GAs). In all the experiments we used a GA with $N = 32$, $\sigma = 32$, population size 256, uniform crossover rate 0.7, uniform mutation and the number of random seeds of 32 (these parameter values were selected to be the same as in \cite{tanese1989distributed}).

We studied search on Type I {\em NM} landscapes with $M=2$ and $P \in \{0,0.1,\dots,1\}$ proportions of all possible second-order interactions.
We studied the search on Type III {\em NM} landscapes with $M \in \{1,3,5\}$ including all possible main effects and odd order interactions of order $\le M$.

\section{Results}
The number of local peaks $L$ in {\em NM} landscapes increases relatively smoothly as we increase the number of terms ($m$) in both Type I and Type II {\em NM} landscapes (i.e., the regions between the vertical lines on Fig. \ref{fig2}) and as we increase the maximum order of interactions $M$ (i.e., as we cross a vertical line on Fig. \ref{fig2}).
\begin{figure}[h]
\hspace{5pt}\includegraphics[width=3.5in]{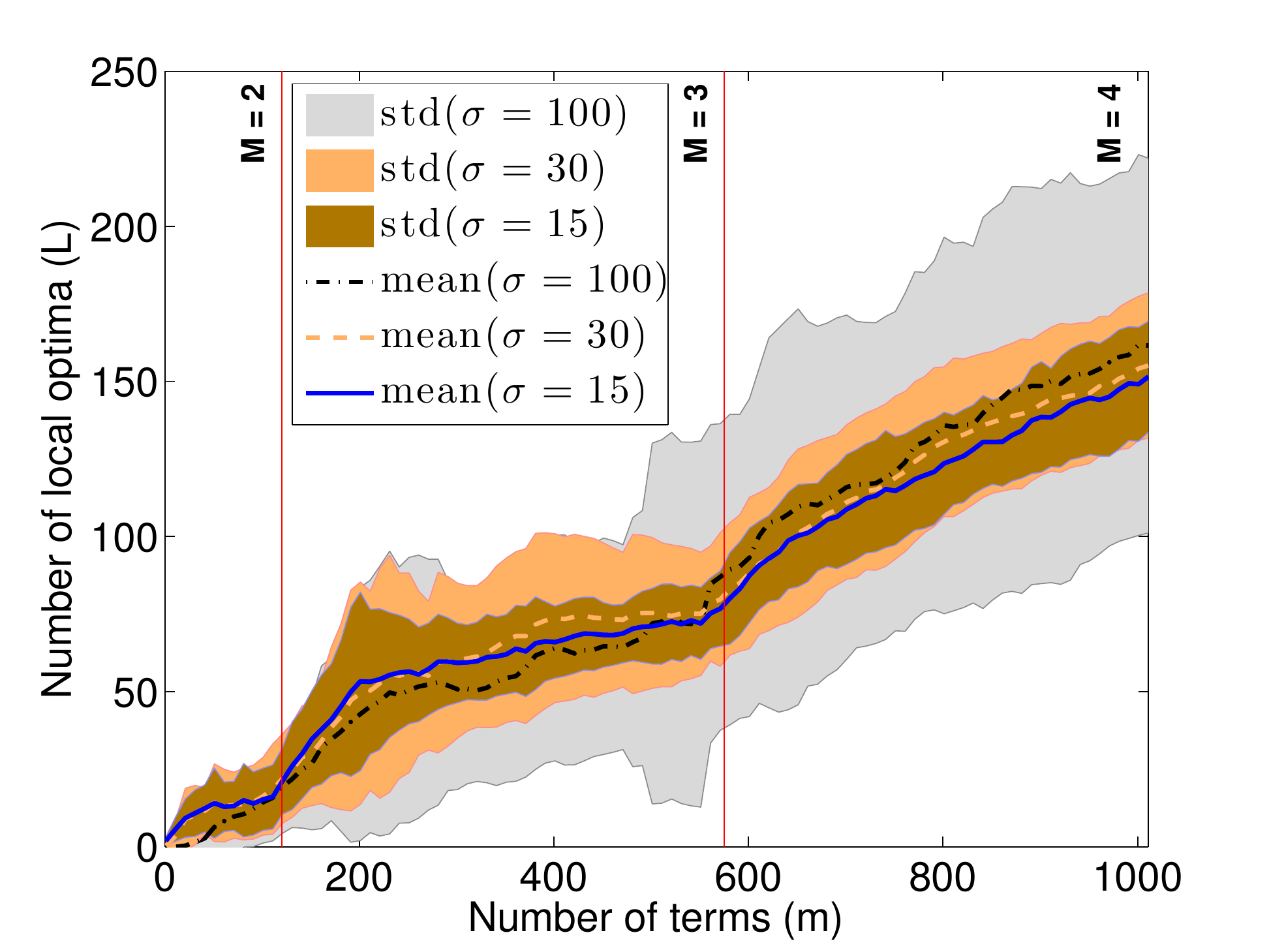}
\caption{Number of local peaks as a function of the number of terms $m$ (x-axis) and order of interactions $M$ (labels near top), for Type II {\em NM} landscapes with $N=15$ and $\sigma \in \{15,30,100\}$. The shaded areas show one standard deviation and the lines show the means for 100 random Type II {\em NM} landscapes. }
\label{fig15}
\end{figure}

\begin{figure*}[h!]
\hspace{0pt}\includegraphics[width=2.5in]{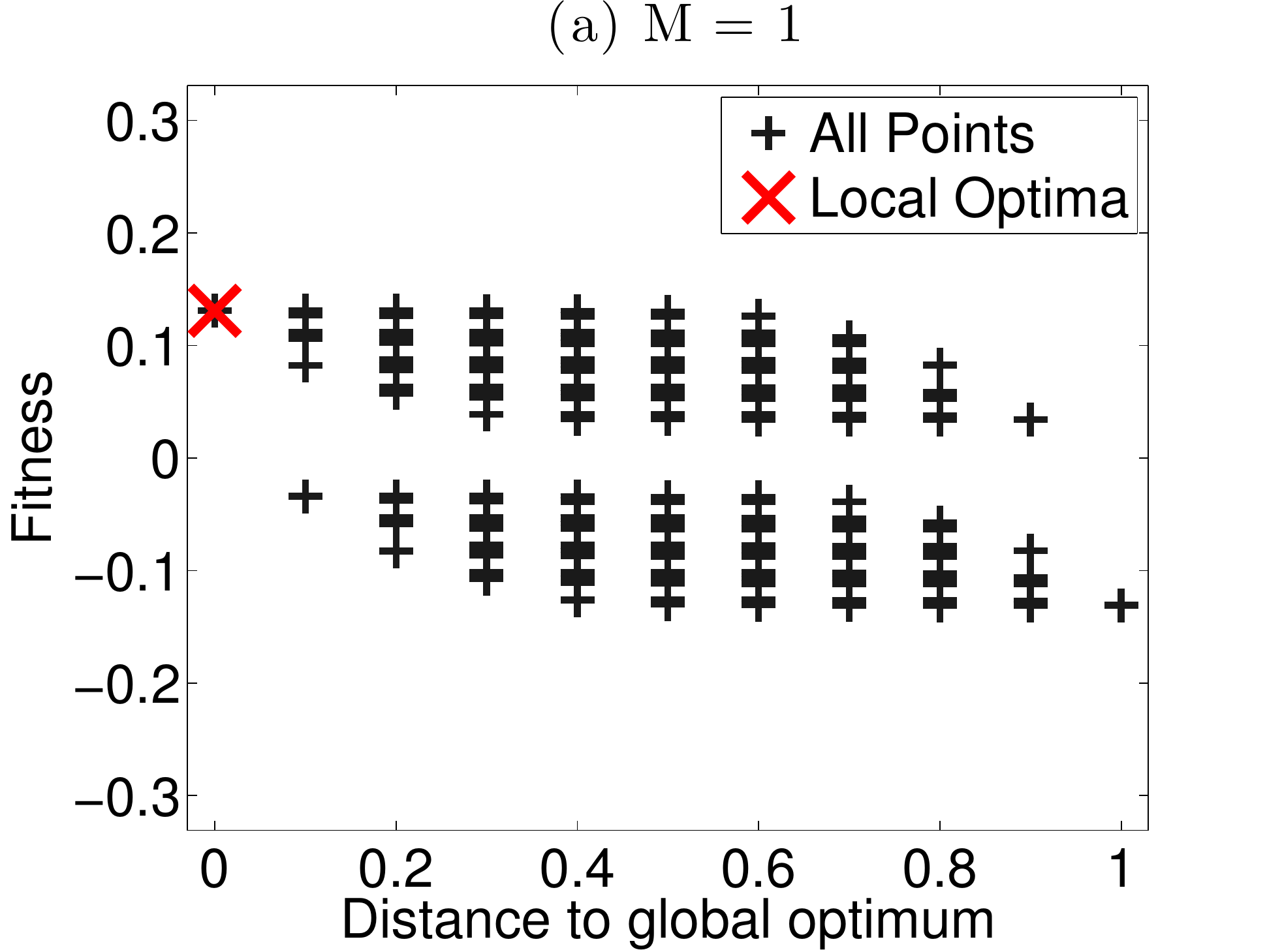}\hspace{-5pt}\includegraphics[width=2.5in]{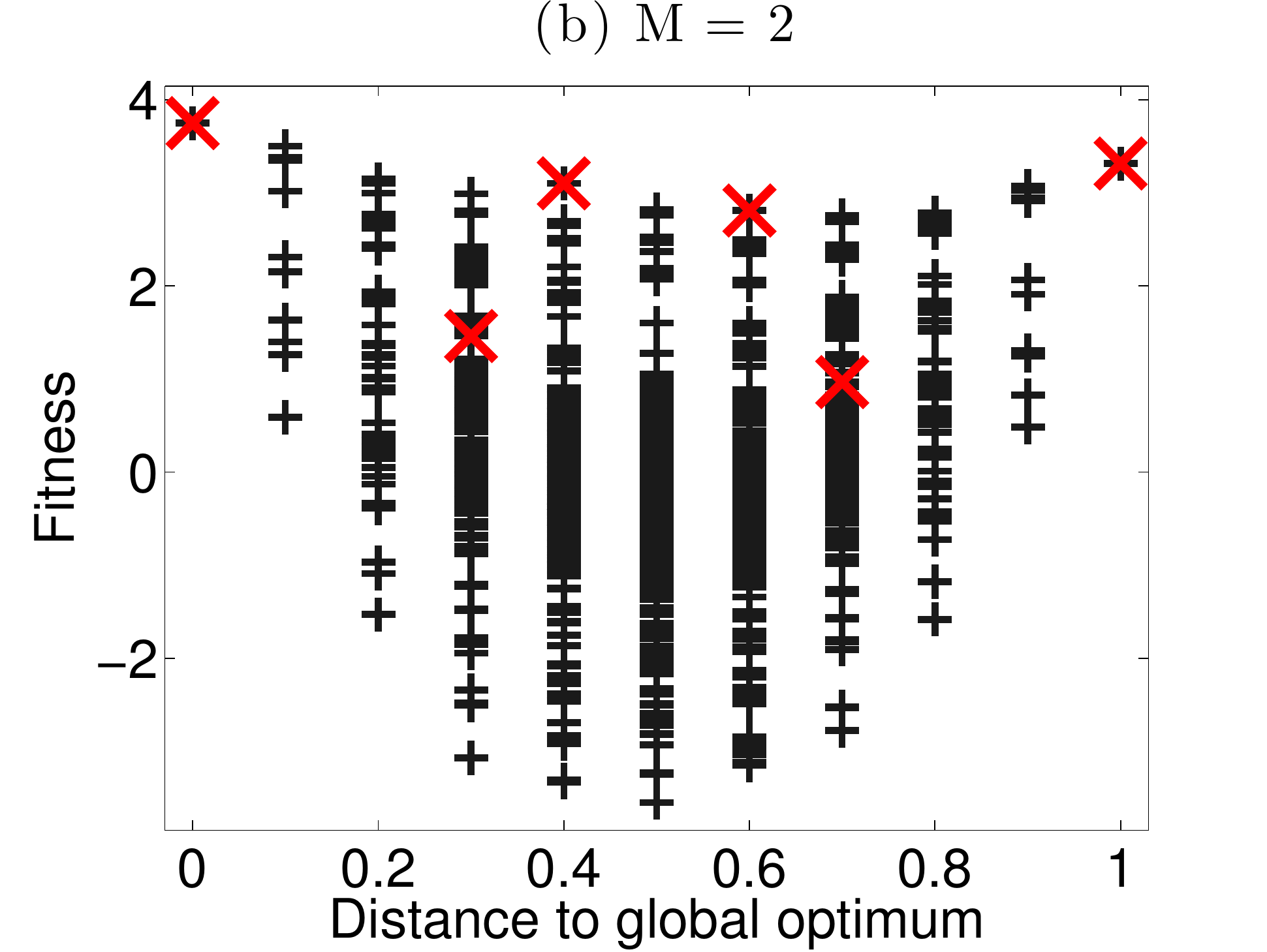}\hspace{-5pt}\includegraphics[width=2.5in]{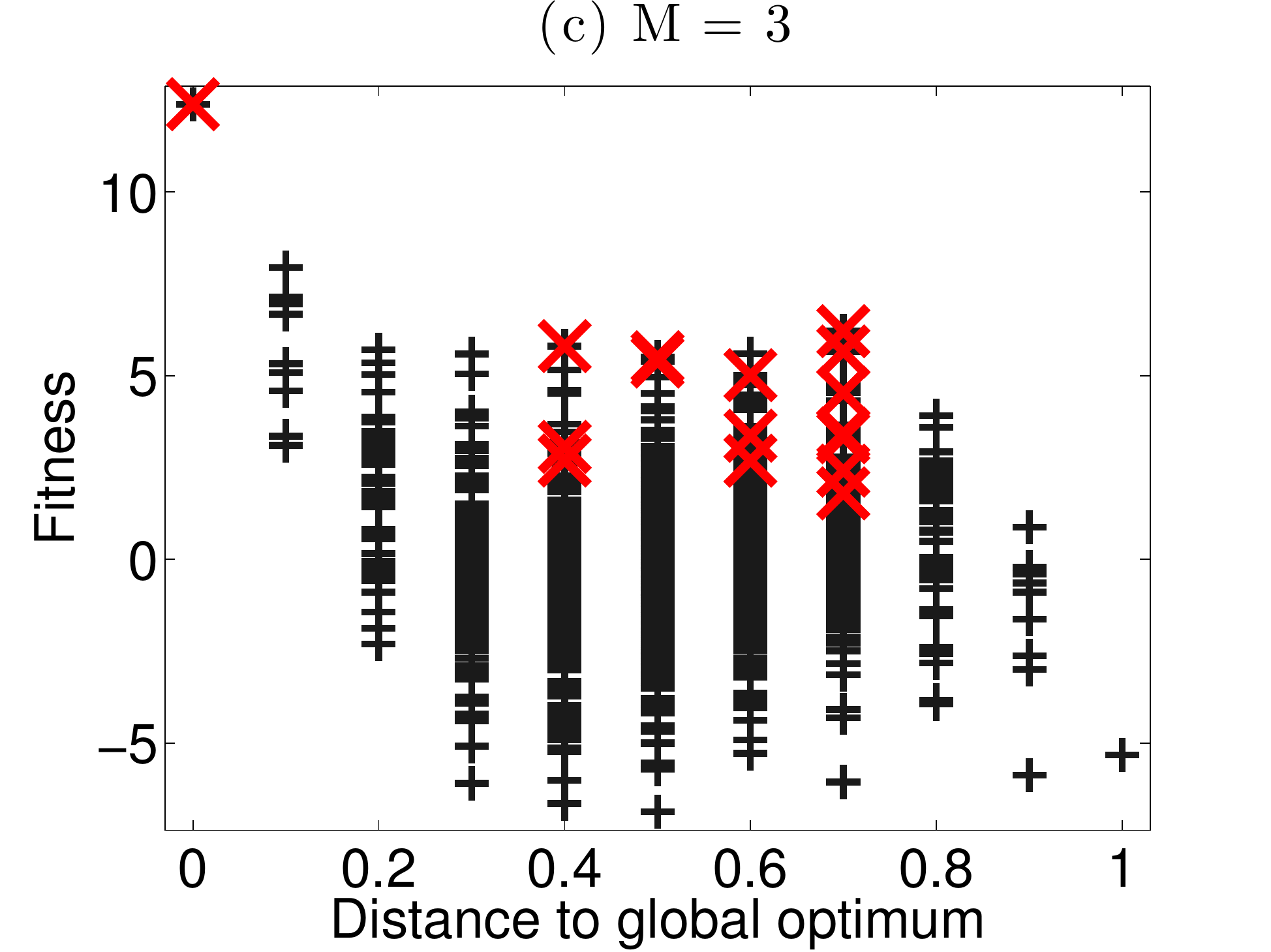}

\vspace{10pt}
\hspace{0pt}\includegraphics[width=2.5in]{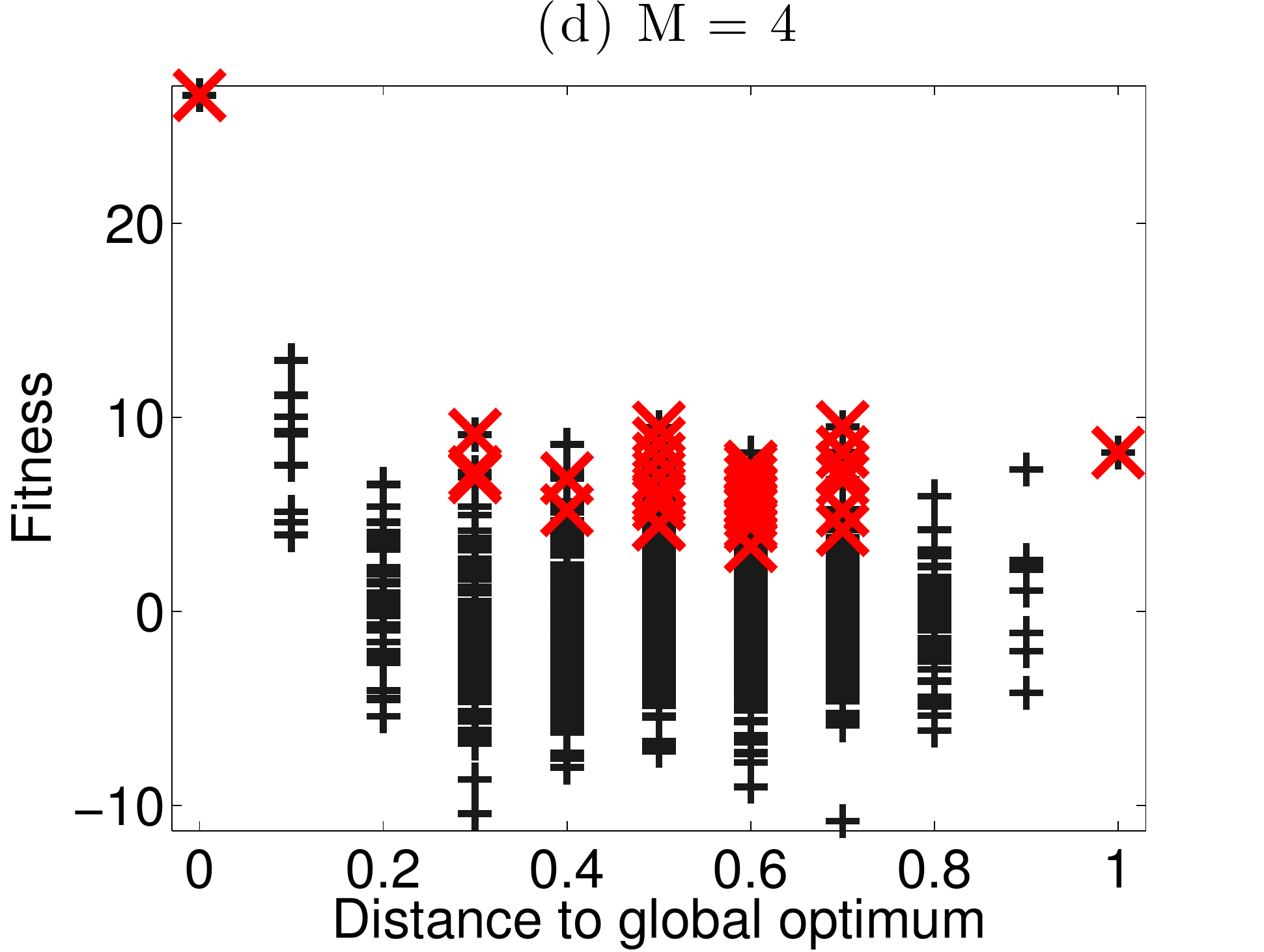}\hspace{-5pt}\includegraphics[width=2.5in]{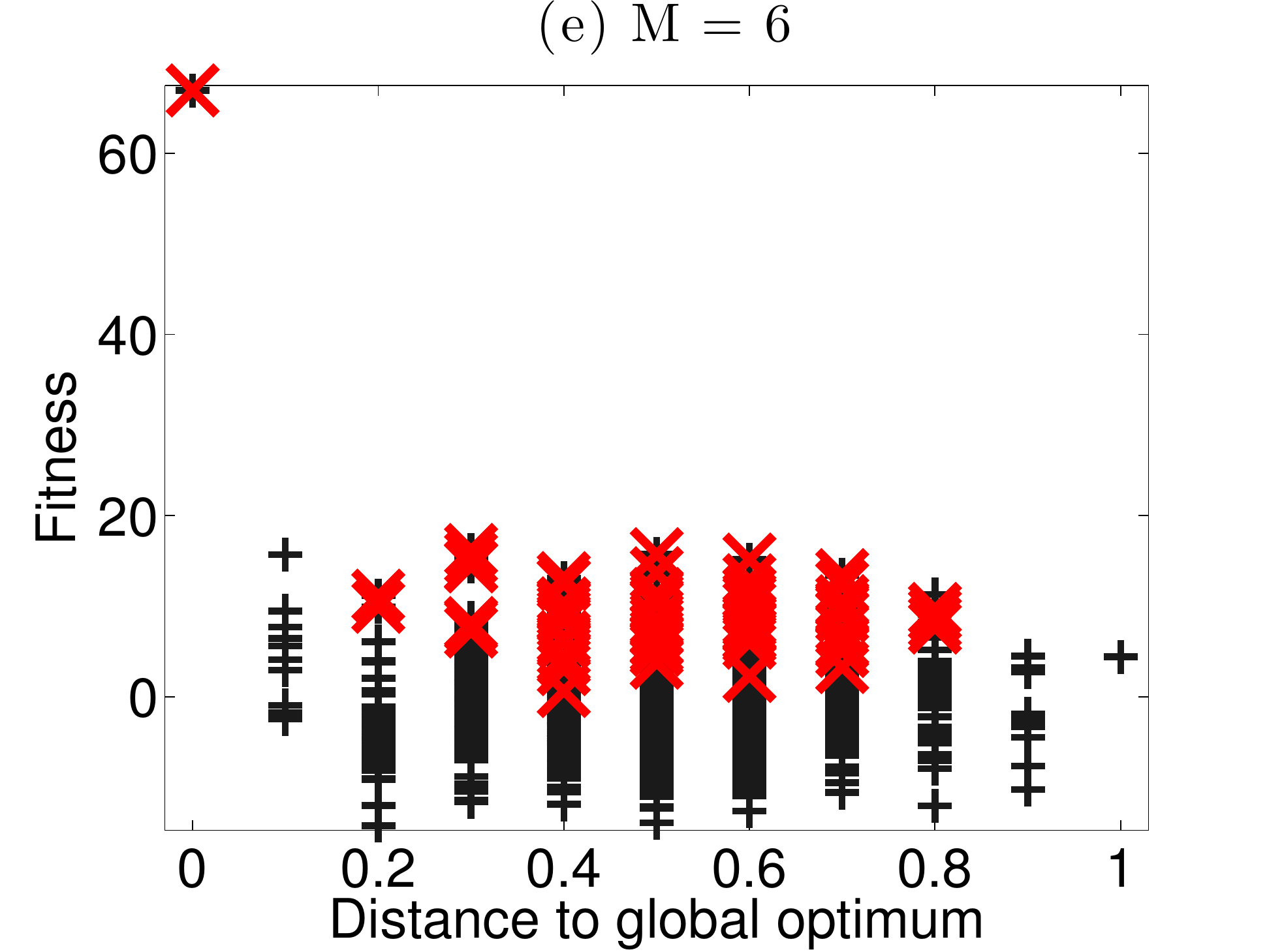}\hspace{-5pt}\includegraphics[width=2.5in]{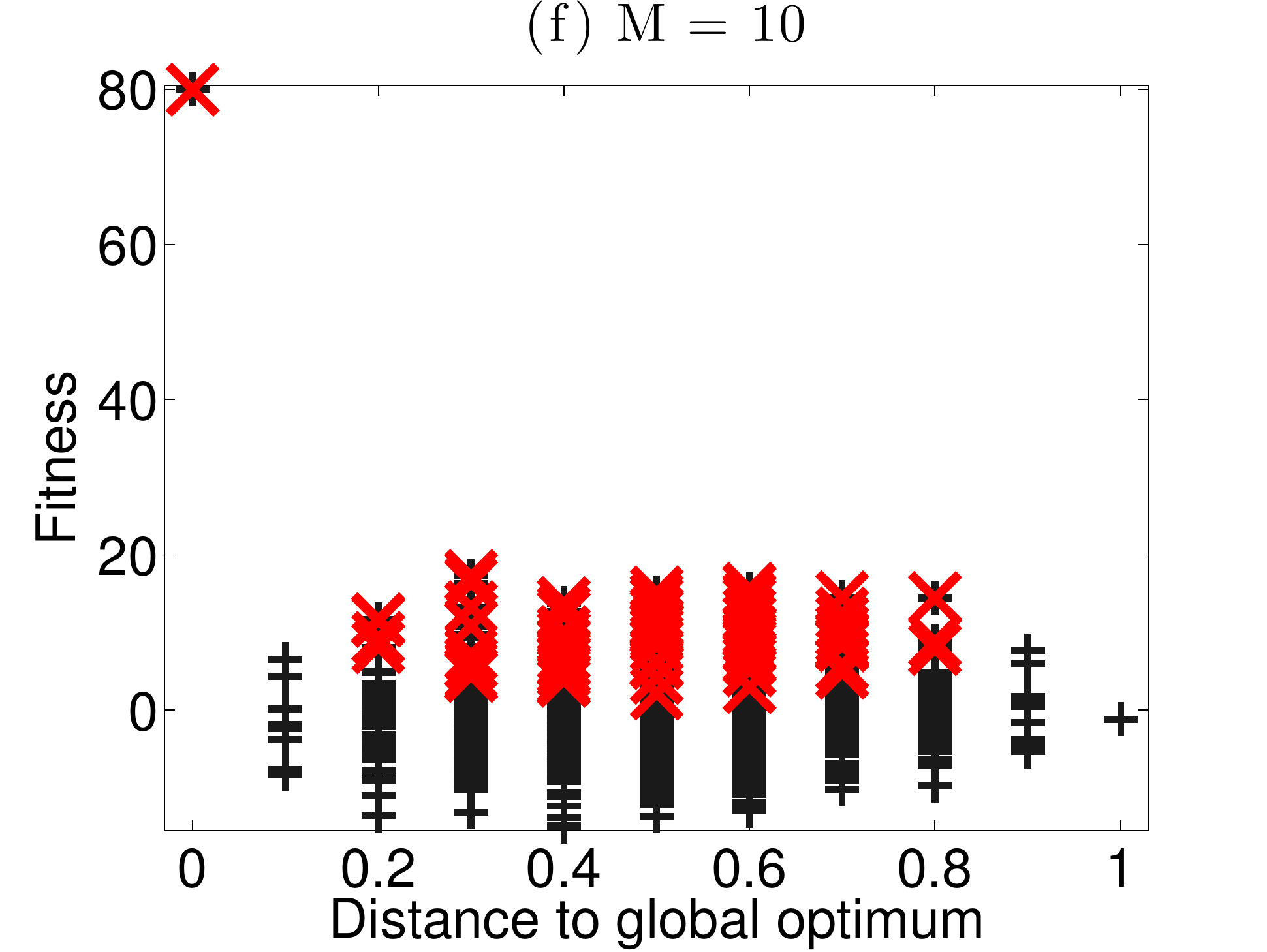}
\caption{Visualization of fitnesses of all the points in representative Type I binary {\em NM} landscapes with $N=10$, $\sigma = 10$ versus their distances from the global optimum in feature space for (a) {\em M} = 1, (b) {\em M} = 2, (c) {\em M} = 3, (d) {\em M} = 4, (e)    {\em M} = 6, (f) {\em M} = 10. In these models, all possible interactions for orders $\le M$ were included.}
\vspace{25pt}
\label{dist}
\end{figure*}

\begin{figure*}[h!]
\hspace{0pt}\includegraphics[width=2.5in]{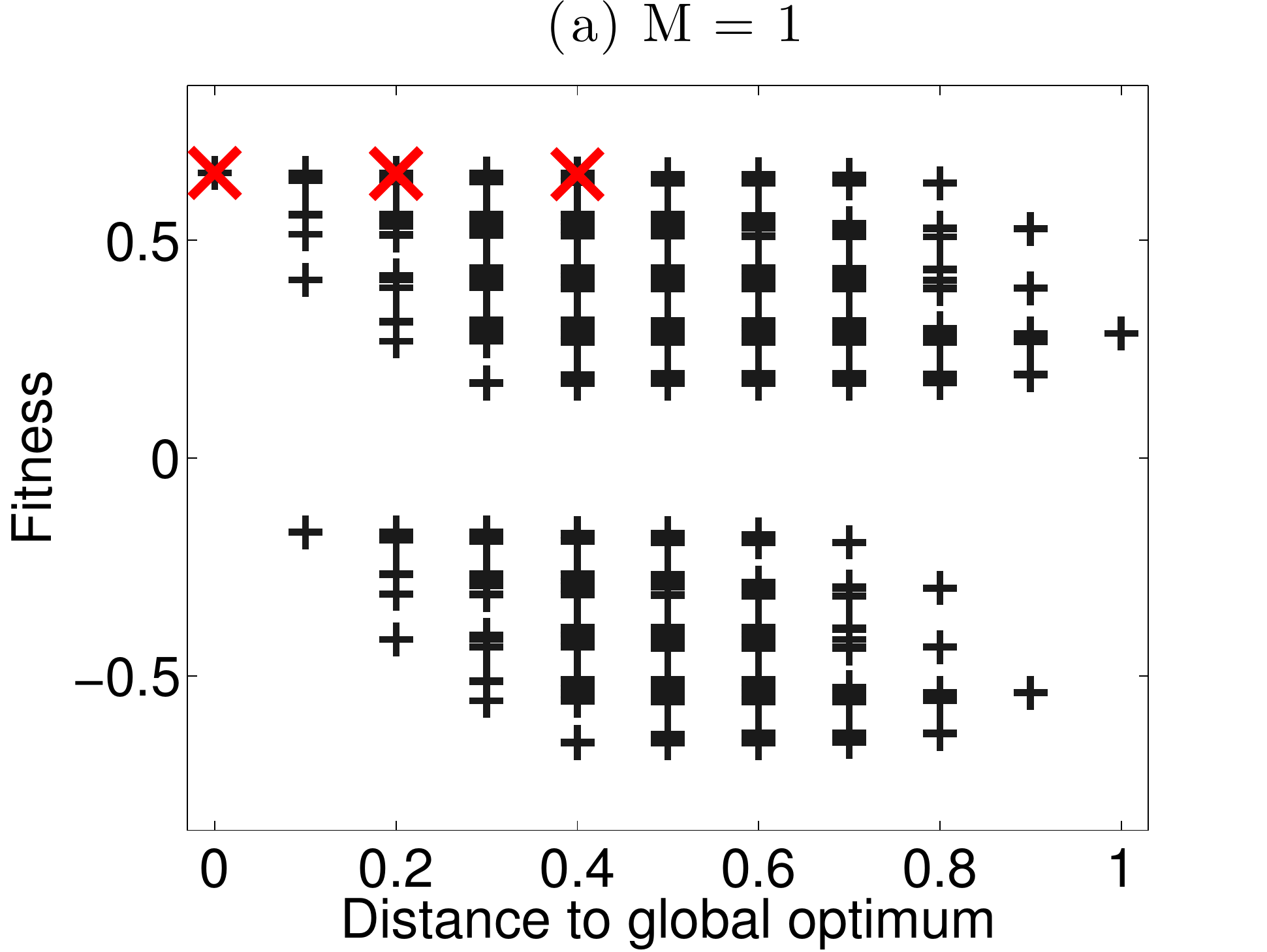}\hspace{-5pt}\includegraphics[width=2.5in]{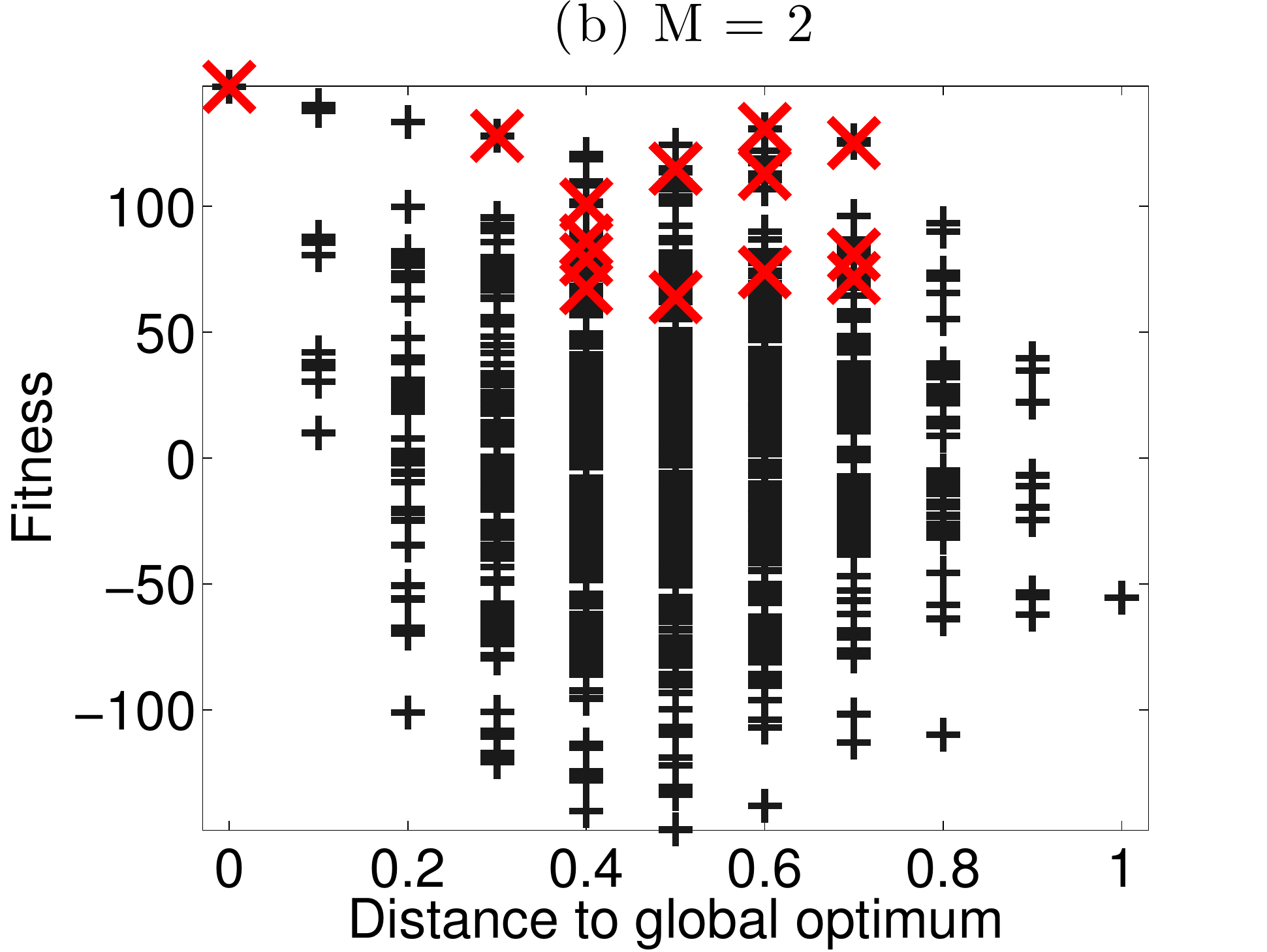}\hspace{-5pt}\includegraphics[width=2.5in]{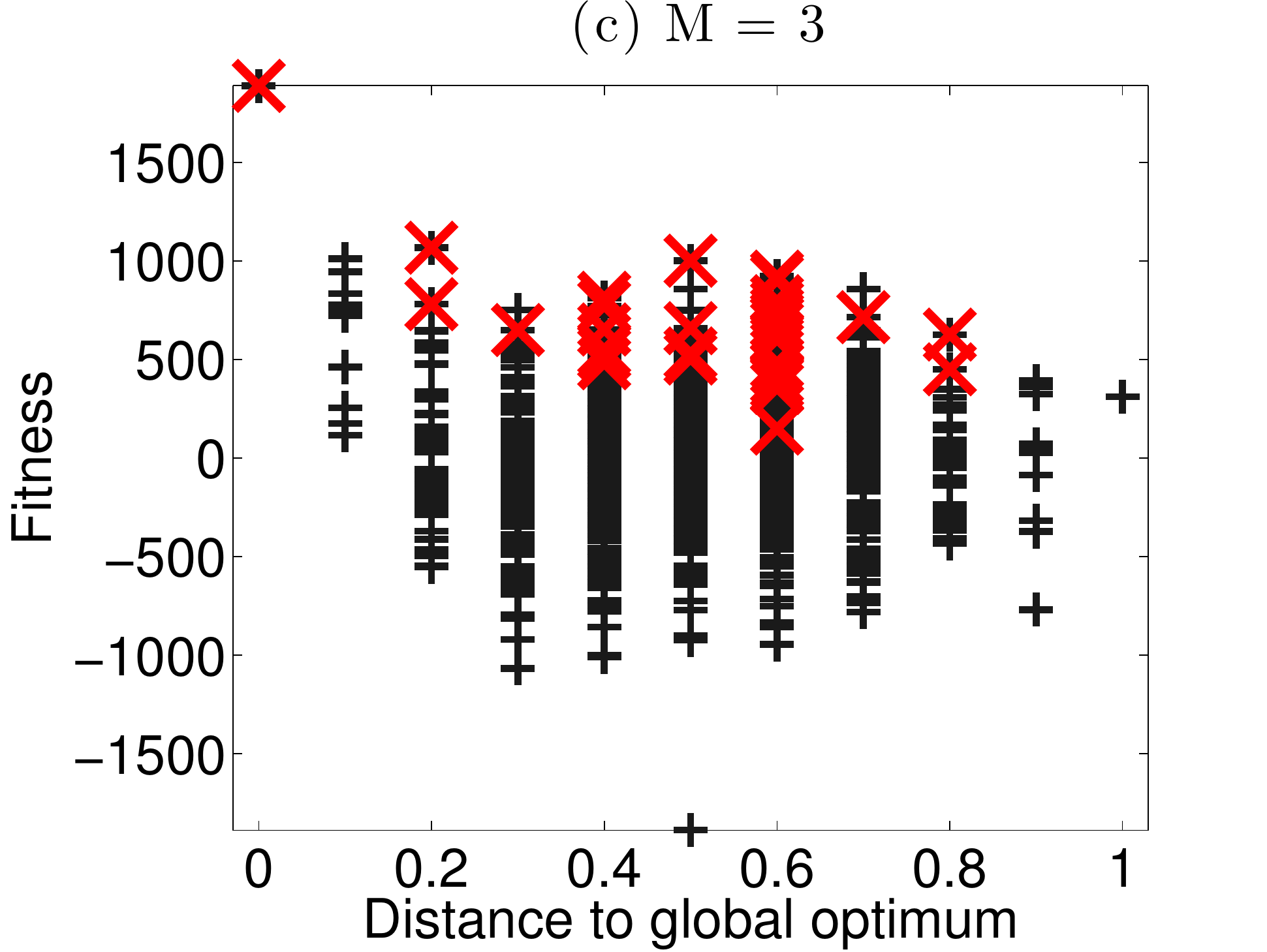}

\vspace{10pt}
\hspace{0pt}\includegraphics[width=2.5in]{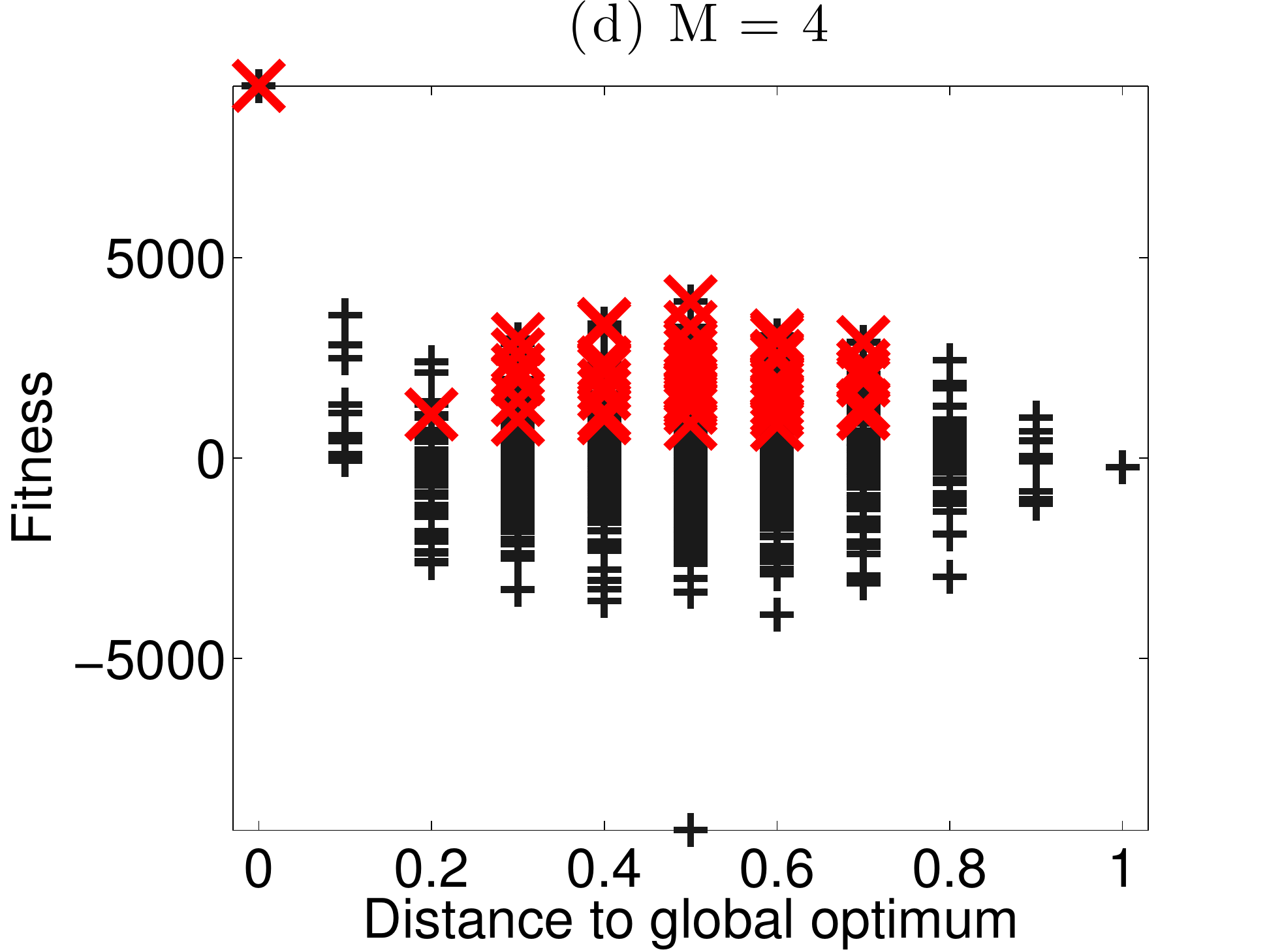}\hspace{-5pt}\includegraphics[width=2.5in]{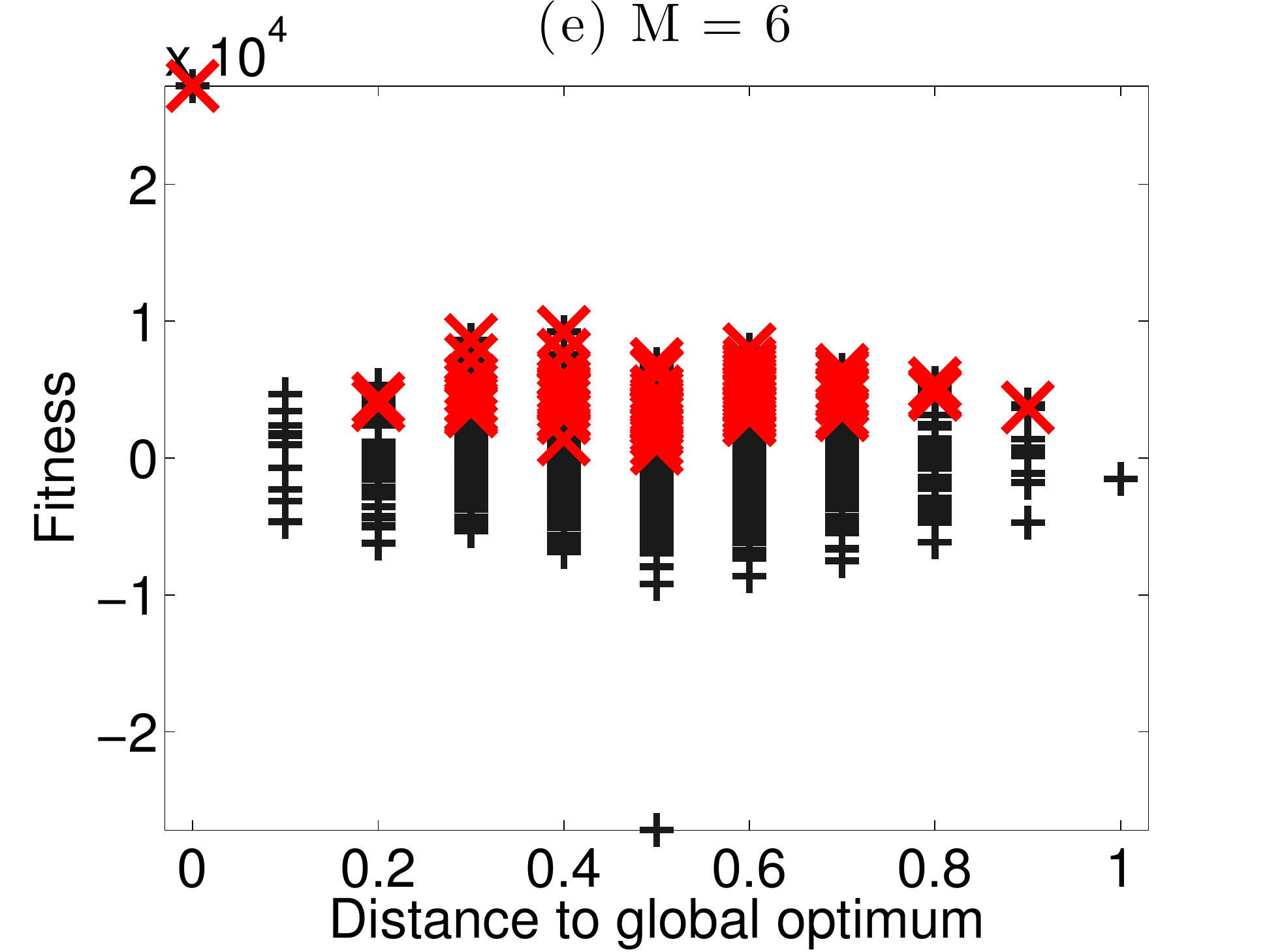}\hspace{-5pt}\includegraphics[width=2.5in]{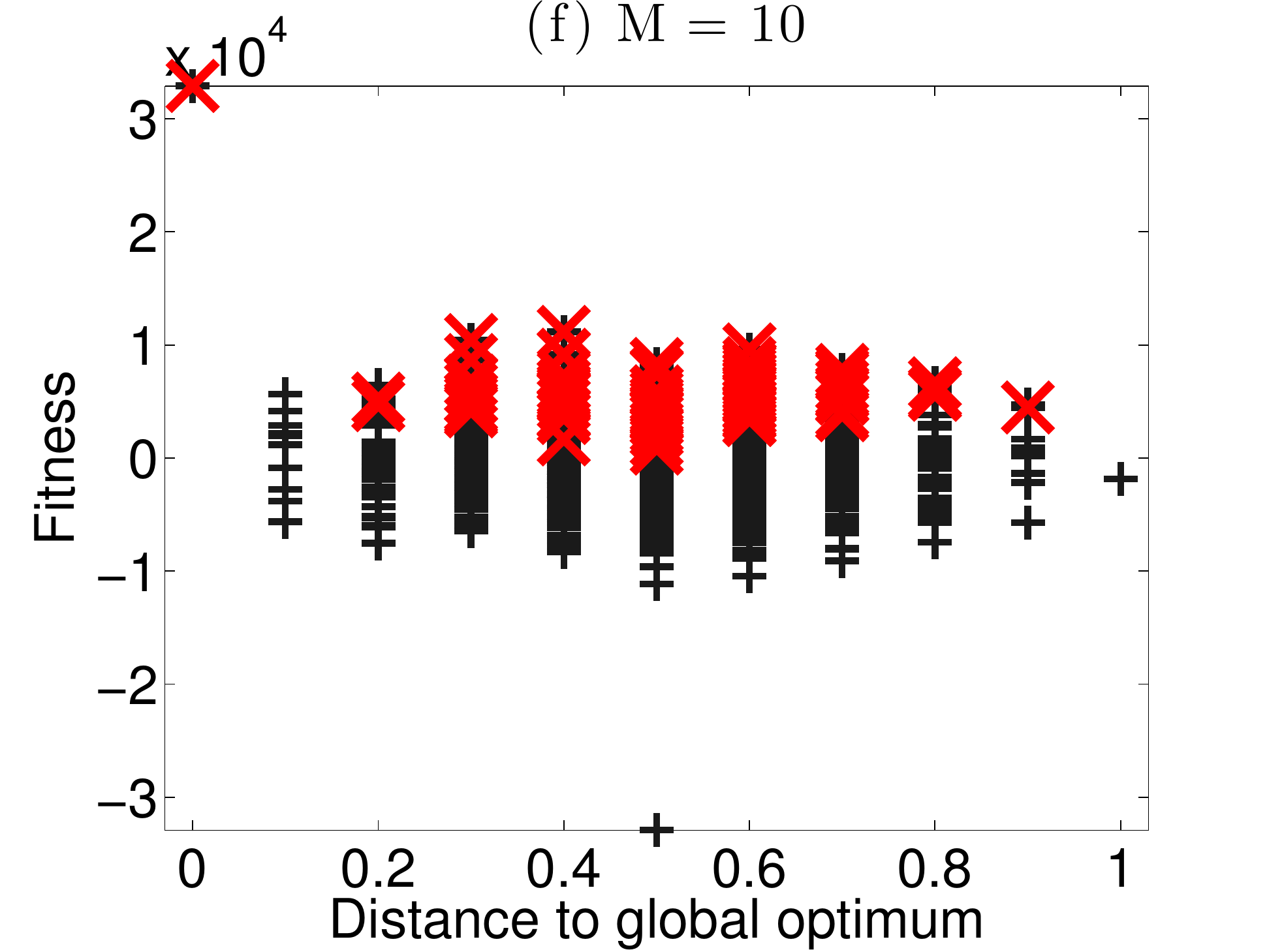}
\caption{Visualization of fitnesses of all the points in representative Type II binary {\em NM} landscapes with $N=10$, $\sigma = 10$ versus their distances from the global optimum in feature space for (a) {\em M} = 1, (b) {\em M} = 2, (c) {\em M} = 3, (d) {\em M} = 4, (e)    {\em M} = 6, (f) {\em M} = 10. In these models, all possible interactions for orders $\le M$ were included.}
\label{distMin}
\end{figure*}

\begin{figure}[h]
\hspace{3pt}\includegraphics[width=3.5in]{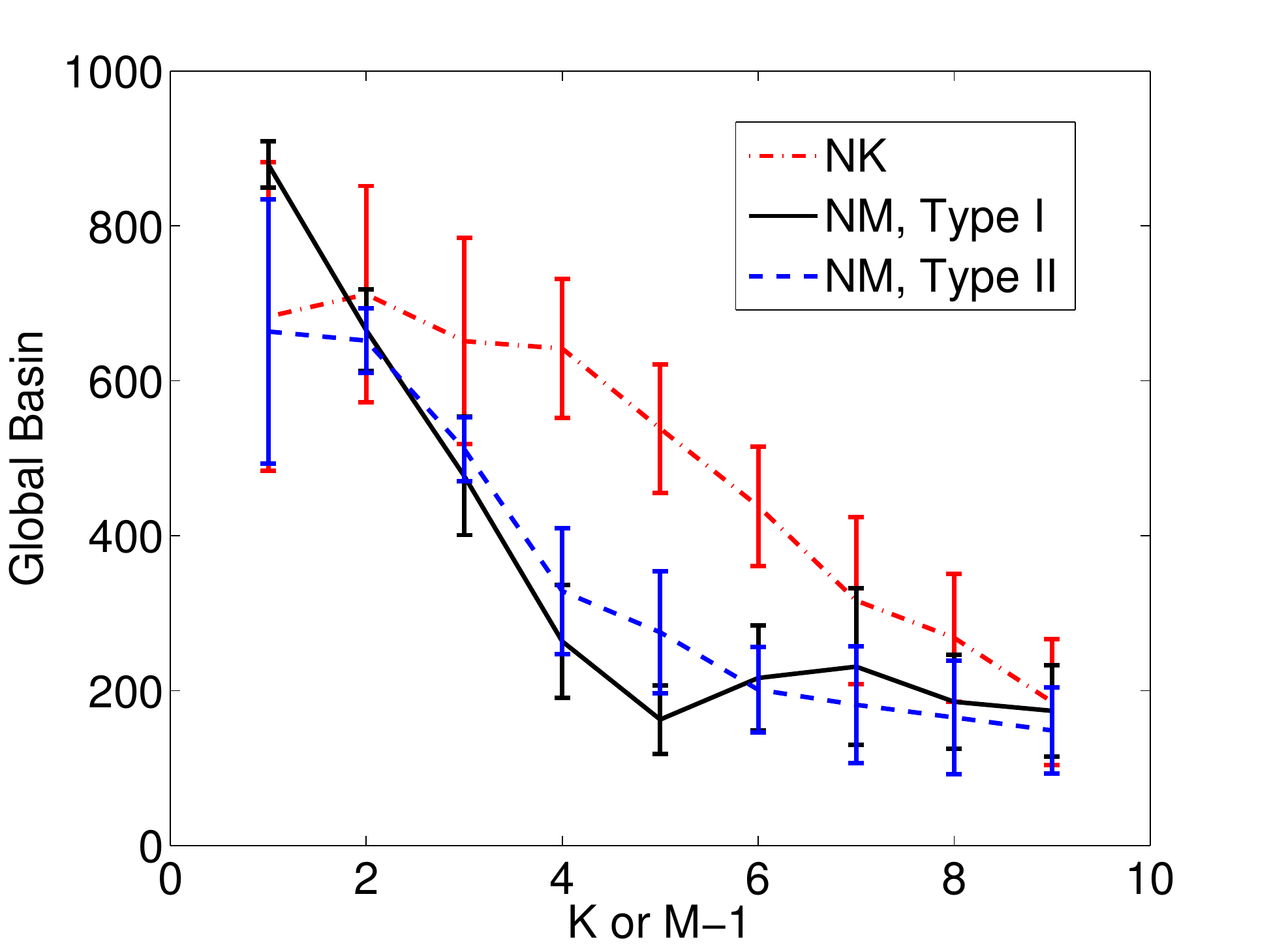}
\caption{The size of the basin of attraction for the global optimum for {\em NK}, Type I and Type II {\em NM} landscapes as a function of the maximum order of the interactions ($K = M-1$).}
\label{figNKNM}
\end{figure}

\begin{figure}[h]
\hspace{3pt}\includegraphics[width=3.5in]{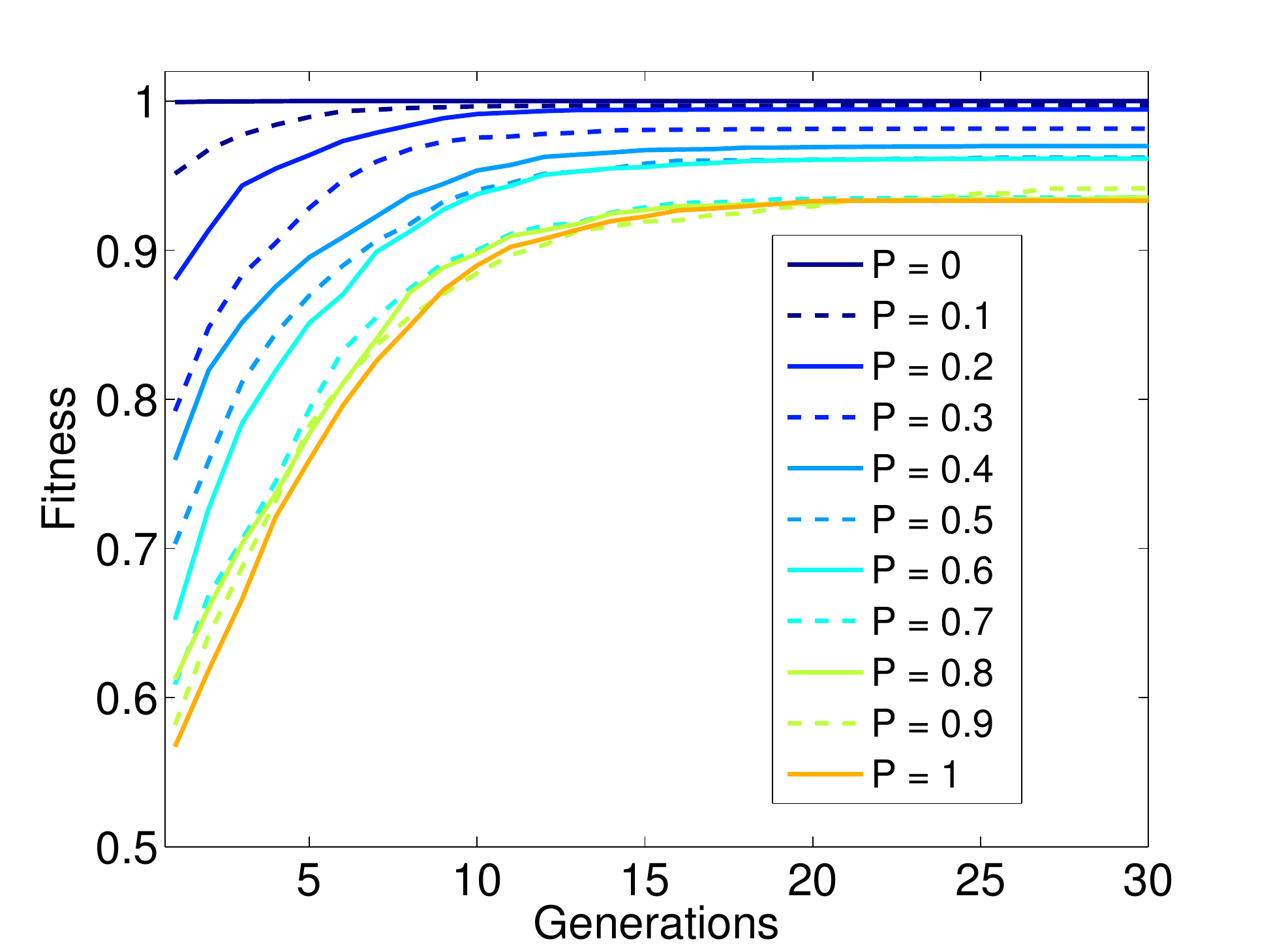}
\caption{Average of the best fitnesses in populations of 256 agents shown for 30 generations of GA search performed on 32 random {\em NM} landscapes with $M = 2$, $N = 32$, and different proportions (P) of second-order interactions.}
\label{searchBest}
\end{figure}

\begin{figure}[h]
\hspace{-10pt}\includegraphics[width=4in]{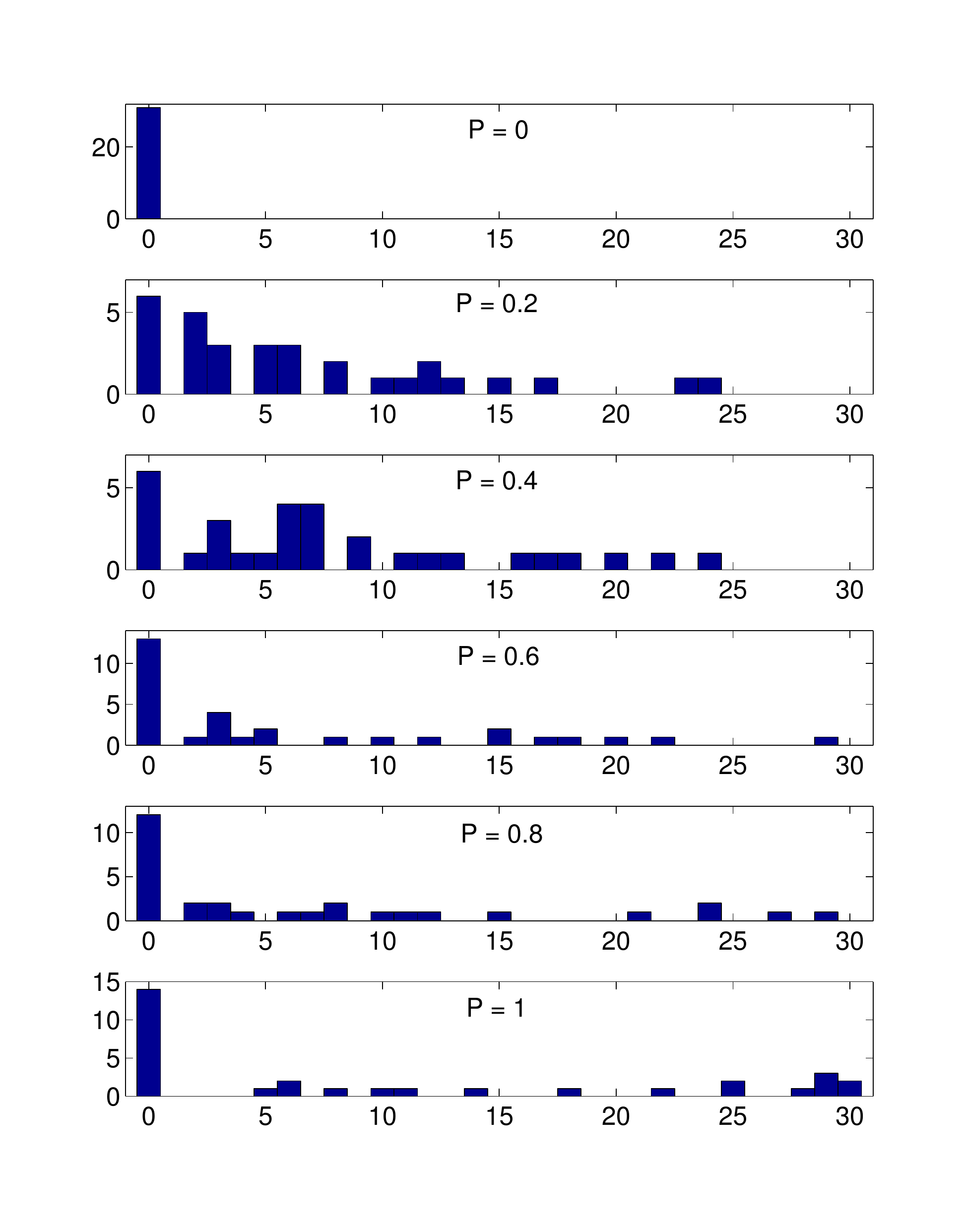}
\vspace{-40pt}
\caption{Histograms of the distances between the best solution at generation 30 and the global maximum for 32 random Type I {\em NM} binary landscapes with $M = 2$ and $N = 32$, and with proportions (P) of second-order interactions from top to bottom, as indicated.}

\label{histNM}
\end{figure}
\begin{figure}[h]
\hspace{3pt}\includegraphics[width=3.5in]{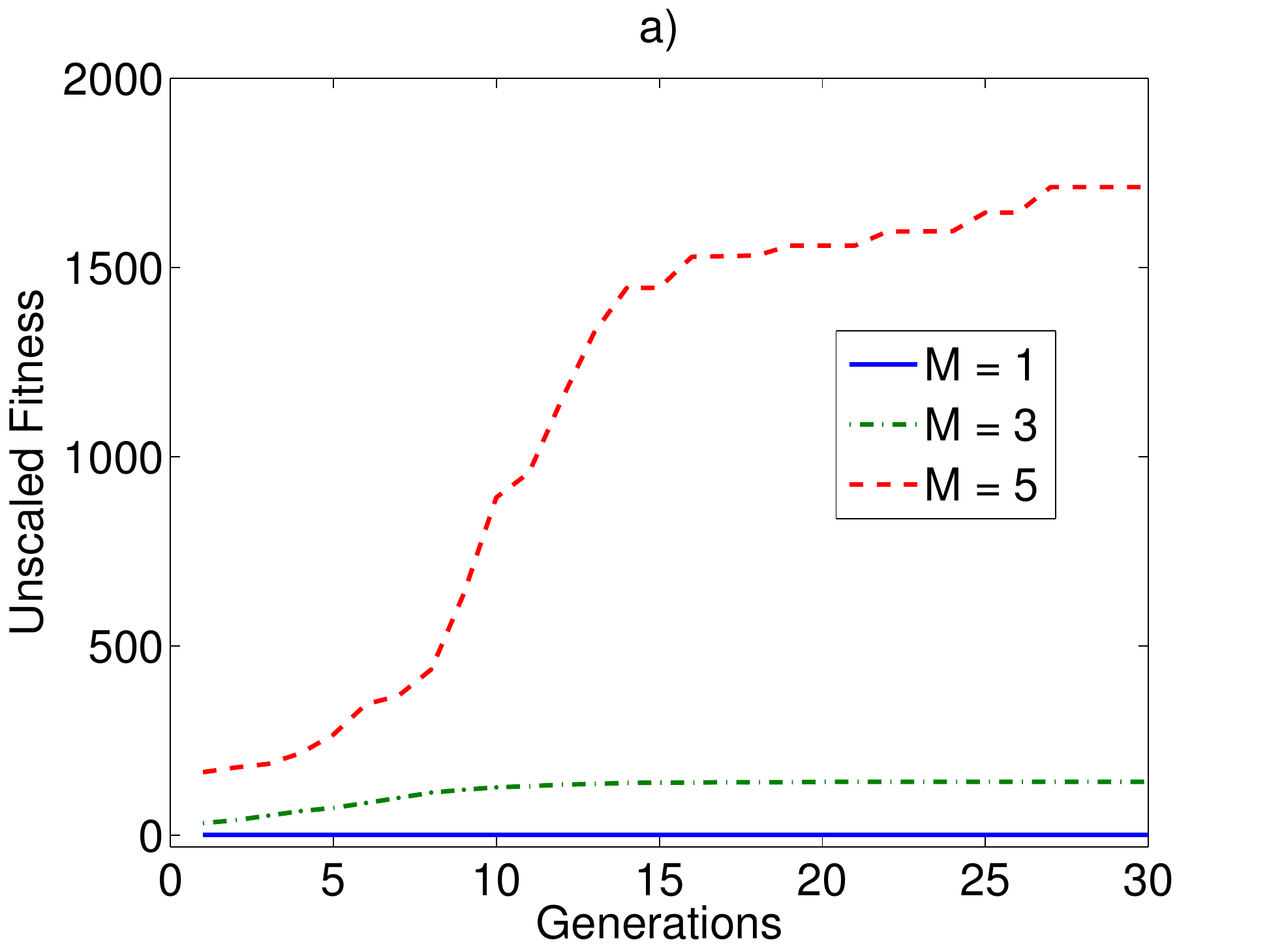}\hspace{-5pt}

\includegraphics[width=3.5in]{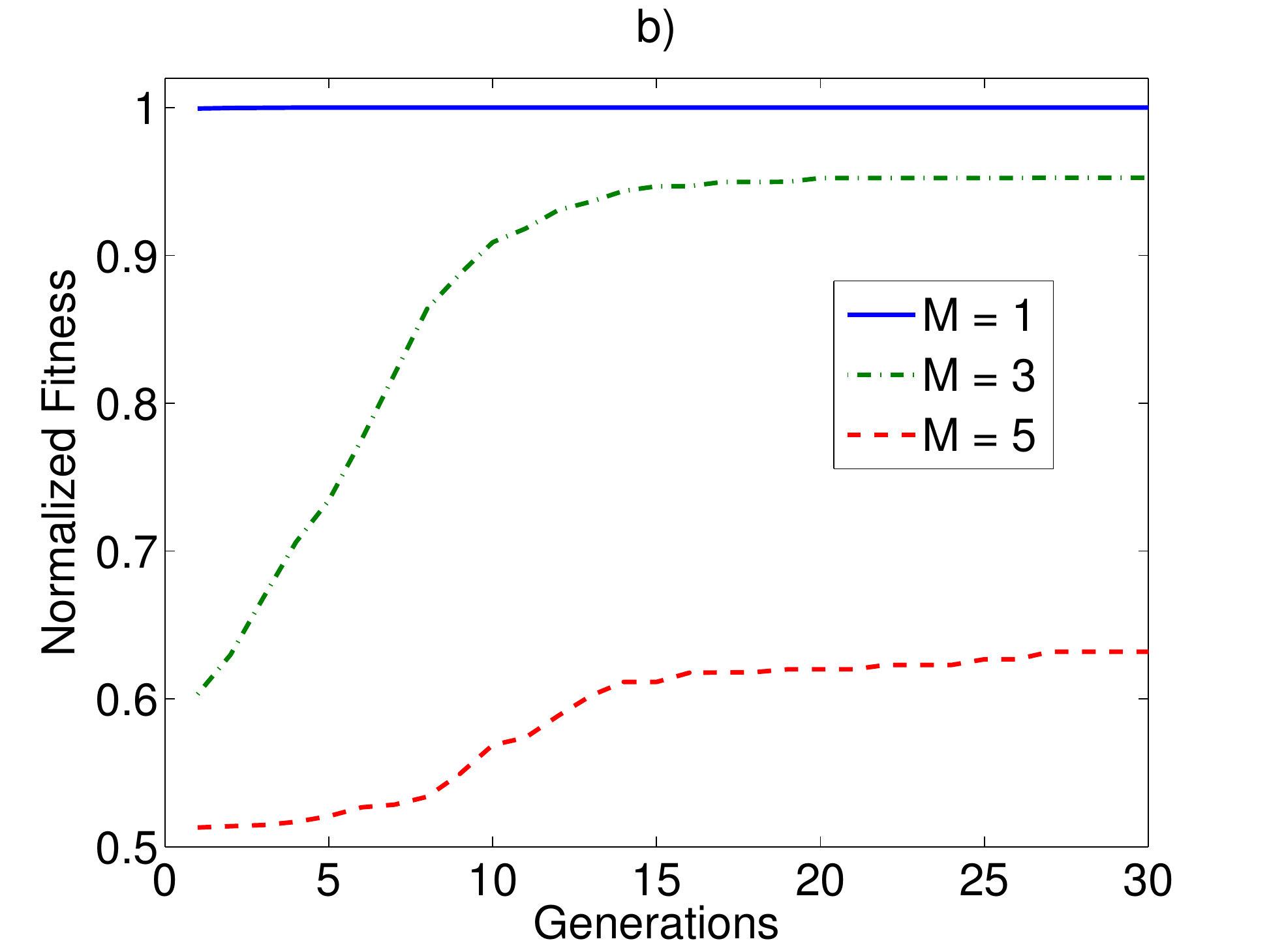}\hspace{3pt}
\caption{Mean of the best fitnesses found by the GA over 30 generations (x-axis) for 32 random Type III  {\em NM} landscapes with $M \in [1, 3, 5]$ and $N = 32$ when (a) fitnesses are not normalized, (b) fitnesses are normalized by Eq. \ref{NORMBYMINMAX}.}
\label{searchBestM1}
\end{figure}

\begin{figure}[h]
\hspace{-1pt}\includegraphics[width=1.8in]{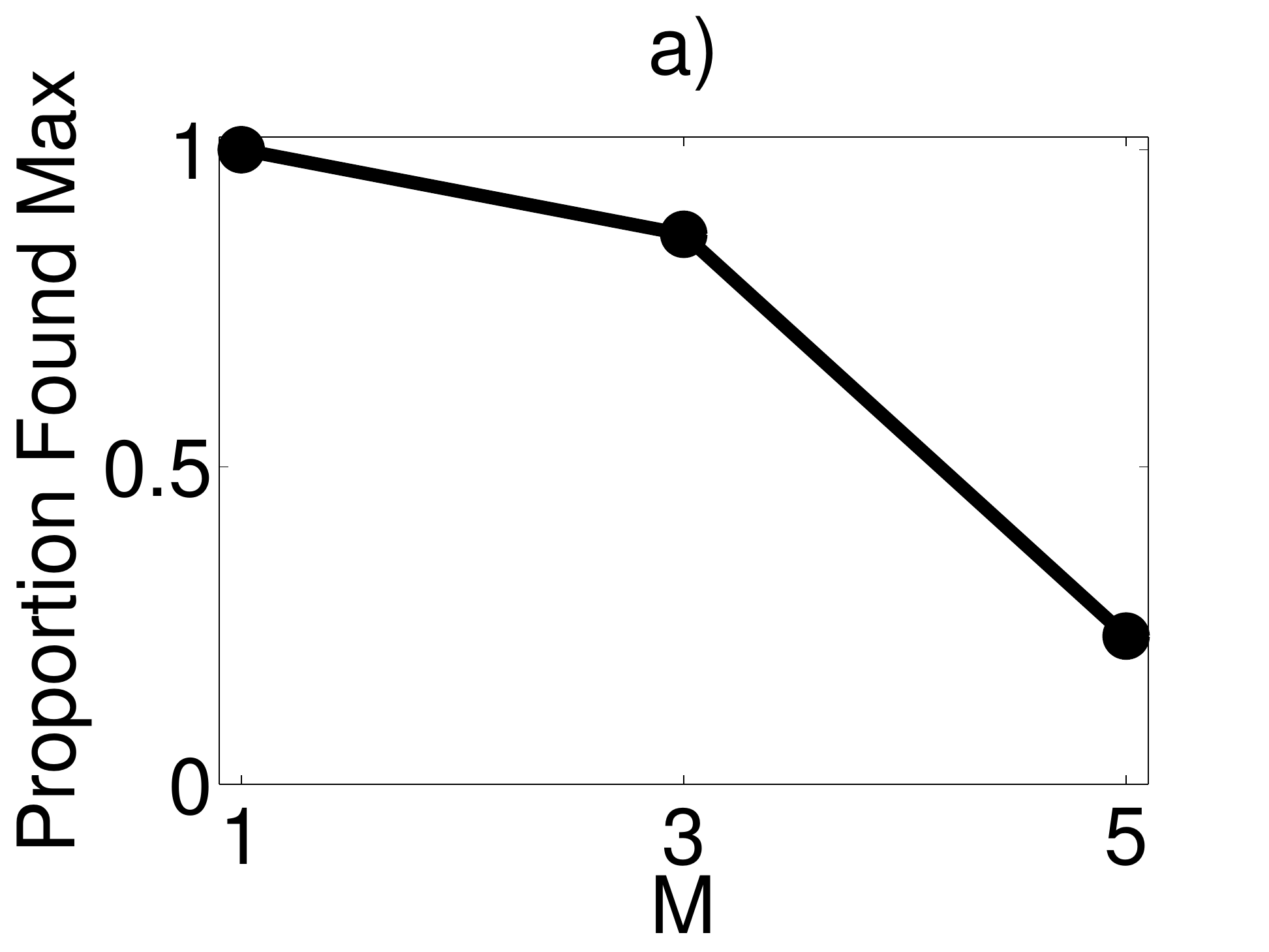}\hspace{-8pt}\includegraphics[width=1.8in]{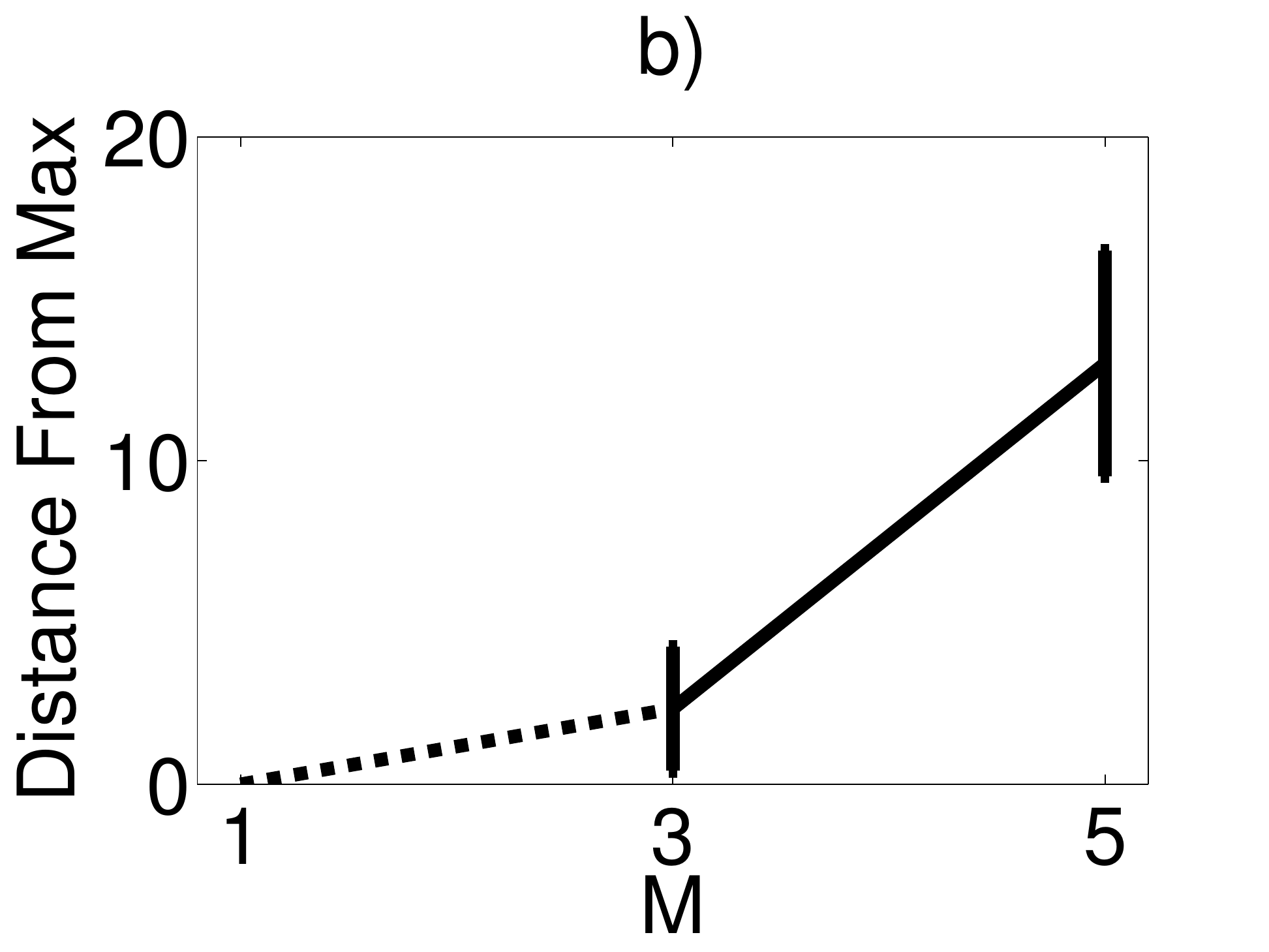}
\caption{Results of search with GA on 32 random {\em NM} landscapes with $M = [1,3,5]$ and $N = 32$. (a) The proportion of the times the search found the global maximum out of 32 runs. (b) The mean and the standard deviation of the distances between the best solution found by GA and the global maximum, when the global maximum was not found. The dashed line indicates that at $M=1$ all runs found the global maximum.}
\label{searchMM}
\end{figure}

\begin{figure}[h]
\hspace{0pt}\includegraphics[width=3.5in]{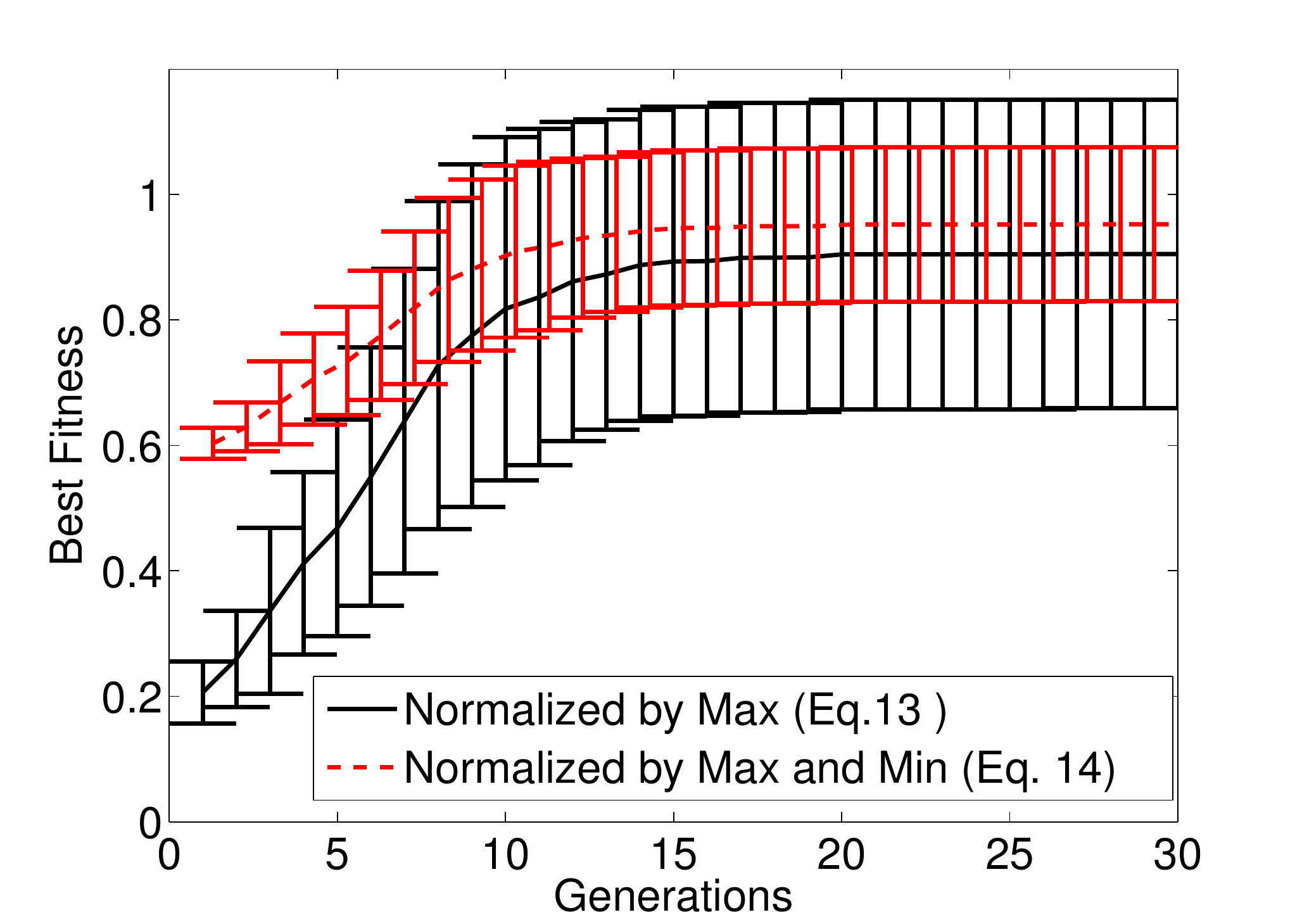}
\caption{The mean and the standard deviation of the best fitnesses found by GA on {\em NM} landscapes with $M = 3$ and $N = 32$, when fitnesses are either normalized by Eq. \ref{NORMBYMAX} (red dashed lines) or by Eq. \ref{NORMBYMINMAX} (black lines).}
\label{errorbarMinMax}
\end{figure}

Note that the average number of local peaks for a given $m$ in both Type I and Type III landscapes is on the same order of magnitude as the expected number of local peaks in {\em NK} models with the same $N$ and $K+1=M$ (compare Figs. \ref{fig2}a and \ref{fig2}c to Fig. 1).

Similarly, the lag 1 autocorrelation of random walks through both Type I and Type III {\em NM} landscapes decreases relatively smoothly as the number of terms $m$ is increased in models with a given $M$, as well as when the maximum order of interactions $M$ is increased (Figs. \ref{fig2}b and \ref{fig2}d), where lower autocorrelation corresponds to greater ruggedness. Notice that, especially for small $M$, the increase in ruggedness  (as measured by both the number of local peaks and the lag 1 autocorrelation) asymptotically slows as the number of terms $m$ increases (Fig. \ref{fig2}).

The larger the $\sigma$ values in Eq. (\ref{betadist}), the smaller the range of fitness values in the landscape (Fig. 2), resulting in larger standard deviations of both the number of local peaks in the landscape (shown in Fig. \ref{fig15} for $N = 15$) and the autocorrelation (not shown).

We show the fitnesses of all points in representative {\em NM} landscapes with $N = 10$ and $\sigma = 10$  as a function of their distances in feature space to the global maximum for both the Type I (Fig.\ref{dist}) and Type II (Fig. \ref{distMin}) {\em NM} landscapes. The global maximum is indicated by the leftmost red $\times$ in each panel and the remaining red $\times$'s are sub-optimal local peaks. As we increase the maximum order of interactions $M$, the fitness difference between the global maximum and other points in the landscape increases; this effect is amplified for Type II {\em NM} landscapes (Fig. \ref{distMin}) relative to Type I {\em NM} landscapes (Fig. \ref{dist}). 

In both Type I and Type II {\em NM} landscapes the distance in feature space between the global maximum and the nearest local peak generally decreases with increasing $M$ and the sizes of the basin of attraction for the global maximum decreases (Fig. \ref{figNKNM}). Our results show that {\em NK} and {\em NM} landscapes have similar sizes of the basin of attraction for the global maximum for small and large $K$ = ($M-1$). However, the size of the basin of attraction for the global maximum of both Type I and Type II {\em NM} landscapes decreases with increasing $M$ rapidly for $M \le 5$ then levels out,  while for {\em NK} landscapes the decrease is more gradual (see Fig. \ref{figNKNM}). 

The difficulty of GA search on {\em NM} landscapes also increases with increasing $m$ and $M$, by several measures of difficulty. When the maximum order of interactions $M = 2$ and the proportion $P$ of all the possible second-order interactions increases from 0 to 1 in 0.1 increments, our results show that the mean of the best fitnesses found by the GA decreases, although above $P =0.7$ there is little if any further change in difficulty (Fig. \ref{searchBest}). We speculate that this might correspond to the periodically asymptotic pattern in the ruggedness noted previously as the maximum number of terms $m$ approaches the maximum possible for a given $M$ (Fig. \ref{fig2}a-b).  Results are shown for only the first 30 generations, after which no further improvement was observed. Histograms of the Hamming distances between the best solutions found by GA and the global maximum  are shown for 32 runs of the GA, for different proportions of the possible second-order interactions (Fig. \ref{histNM}). 
For unimodal landscapes ($M = 1$), the GA found the global optimum in all 32 runs (Fig. \ref{histNM}). For more rugged landscapes the global optimum was also found in some runs, and surprisingly the proportion of times it was found increased from $P = 0.2$ to $P= 1$. However, as the ruggedness increased, those runs in which the best individuals were suboptimal generally became stuck farther and farther from the global optimum (note how the distributions become increasingly spread out to the right, as you view the panels in Fig. \ref{histNM} from top to bottom).

When fitnesses are not normalized,  a higher maximum order of interactions $M$ results in higher raw fitnesses (Fig. \ref{searchBestM1}a). This is due to the fact that summing more interaction terms result in higher ranges of fitness (Fig. \ref{dist} and Fig. \ref{distMin}). However, when fitnesses are properly normalized by Eq. \ref{NORMBYMINMAX} to the range [0,1], increasing the maximum order of interactions in the model decreases the values of the best individuals' fitnesses found (Fig. \ref{searchBestM1}b), reflecting the fact that GA search becomes more difficult at higher $M$. 

While the proportion of times that GA was able to find the global maximum out of 32 runs decreased as the maximum order of interactions $M$ increased (Fig. \ref{searchMM}a), the means and standard deviations of the distances between the best solutions found by the GA and the global maximum increased  (Fig. \ref{searchMM}b).
Normalizing by Eq. (\ref{NORMBYMAX}), rather than Eq. (\ref{NORMBYMINMAX}) exaggerates both the apparent relative generational increase in fitnesses in the GA and the variance in fitnesses across different random landscapes with the same $m$ and $M$ (Fig. \ref{searchBestM1}). This illustrates how knowing the global minimum can help to accurately assess the mean and standard deviation of the relative increase in fitnesses and fairly compare search results on different {\em NM} landscapes.

\section{Discussion}
In this work we introduce {\em NM} landscapes, which are parametric interaction models that (a) have non-negative coefficients and (b) are well-defined for feature alphabets of any arity (from binary to real-valued), as long as (c) the minimum value in the alphabet is negative with absolute value less than or equal to the positive maximum. This combination of constraints ensures that a global maximum is located at the point where all decision variables have their maximum value, with the optimal value equal to the sum of the model coefficients.  By further restricting which combinations of interactions are present, various subsets of {\em NM} models also have known location and value of the global minimum (we illustrate two such subsets, which we refer to as Type II and Type III {\em NM} landscapes).  By using an appropriate non-negative distribution for the coefficients, the resulting {\em NM} landscape models have relatively smoothly tunable ruggedness. Epistatic terms are transparently represented in {\em NM} landscapes, making it trivial to control or analyze exactly which terms and interactions are present.  In the following, we discuss why these various aspects of {\em NM} landscapes are valuable, and how they offer advantages over {\em NK} landscapes and Walsh polynomials as epistatic benchmark problems.

\subsection{Value of finely tunable epistasis}
Although {\em NK} landscapes have been widely used as benchmark problems with varying degrees of epistasis, there are many potential applications that require more fine control over which terms are present or absent. 

For example, this study was originally motivated by some of our previous research in comparing search strategies for healthcare improvement \cite{eppstein2012searching,manukyanteam}.  In the context of clinical fitness landscapes, it is not reasonable to assume that all features have only main effects (corresponding to $K=0$ in {\em NK} landscapes) as there are many known interactions between various practices  and/or treatments in the real world (e.g., \cite{johnell2007relationship, dechartres2011single}). However, it is also not reasonable to assume that every feature interacts with at least one other feature (corresponding to $K=1$).  Rather, we sought to explore the performance of different clinical quality improvement strategies (including randomized controlled trials and team quality improvement collaboratives) in more realistic clinical fitness landscapes where all features had main effects but varying numbers of second-order interactions were also present. 
 
Alternatively, in some application domains one may wish to model purely epistatic landscapes in which {\em no} main effects are present. For example, in complex diseases there may be little if any association between single genes and incidence of disease \cite{moore2003}. Similarly, the electrical grid is explicitly ensured to be stable with respect to the loss of any one component, but interactions between two or more component outages can lead to large cascading failures \cite{EppsteinHines}. For these types of  applications, we and others have been seeking algorithms that are capable of detecting purely epistatic interactions (e.g., \cite{eppstein2008, urbanowicz2010, EppsteinHines}). To test these algorithms, one must be able to create benchmark landscapes where there are interaction terms but no main effects. 

Classic {\em NK} landscapes cannot model landscapes between $K=0$ and $K=1$, nor can they model landscapes with no main effects or where the strengths of the main effects are smaller than the strengths of interaction terms \cite{buzas2013analysis}.  In contrast, general interaction models (including {\em NM} landscapes) easily allow fine control over exactly which terms are present or absent and one can easily specify different magnitudes of coefficients for different terms.  This is also possible using Walsh polynomials, although the notation is not as simple or transparent. 

In the experiments shown here, we provide evidence that increasing the number $m$ and/or maximum order $M$ of interactions increases the ruggedness of {\em NM} landscapes with coefficients generated using Eq. (\ref{betadist}) with $\sigma = N$, as measured by number of local peaks and the lag 1 autocorrelation of random walks through the landscapes (Fig. \ref{fig2}), and also increases the difficulty of these landscapes by several different measures of search difficulty, including size of the basin of attraction of the global optimum (Fig. \ref{figNKNM}), final normalized best fitnesses found with a GA (Figs. \ref{searchBest} and \ref{searchBestM1}), distances from the global optimum of sub-optimal final best fitnesses  found by a GA (Figs. \ref{histNM} and \ref{searchMM}a), and proportion of times a GA was able to find the global optimum (Fig. \ref{searchMM}b).

\subsection{Value of fitness normalization}
Since the range of possible fitness values varies so much between rugged landscapes (as illustrated in Figs. \ref{dist} and \ref{distMin}), it is important to normalize fitnesses to a consistent range if one desires to compare fitness values on different lanscapes (Fig \ref{searchBestM1}), or to assess the variability of a search strategy on landscapes with a given $m$ and $M$ (Fig. \ref{searchBest}).  In \cite{eppstein2012searching,manukyanteam} we used logistic transforms of general parametric interaction models with unknown maxima to model search on clinical fitness lanscapes with varying numbers of second order interactions. While the logistic function successfully bounds the transformed fitnesses to the open interval $(0,1)$, it also has the side effect of compressing high fitness values to the degree that there is very little difference between the fitnesses of the optimal peak and many suboptimal peaks. This may be a realistic assumption in health care (where there may be many possible combinations of clinical practices that yield good results), but for applications where such compression is not ideal it may be more appropriate to normalize fitnesses to values $\le 1$ using Eq. (\ref{NORMBYMAX}), which requires knowing the global maximum, or even better to the closed interval $[0,1]$ using Eq. (\ref{NORMBYMINMAX}), which also requires knowing the global minimum. {\em NM} landscapes enable these types of normalization, as discussed in the following subsections.

\subsection{Value of knowing the global maximum}
Knowing the best possible fitness offers obvious benefits, including: (a) one can terminate a search as soon as the known optimal value is found, potentially saving significant computation time; (b) one can compare methods by assessing the frequency with which the search strategies are able to find the global maximum; (c) one can tell if a stalled search has found the global optimum or is stuck on a local optimum; and (d) one can normalize fitnesses to be $\le 1$ using equation (\ref{NORMBYMAX}).
Knowing the location of the global maximum in feature space offers obvious additional benefits \cite{jones1995fitness} including: (e) one can track the evolving distances of solutions to the global optimum as the search progresses, which could potentially inform ways to improve the search process; (f) one can compare the distances (in feature space) from the best final solution to the global optimum; (g) one can assess the difficulty of the fitness landscape by assessing the correlation of fitness values encountered on a random walk with the distances to the global optimum; and (h) one can empirically explore a landscape near the global optimum in order to asses the size and shape of its basin of attraction.

For arbitrary epistatic landscapes (including {\em NK} landscapes, general parametric interaction models, and Walsh polynomials) finding the global maximum is NP complete. However, there are restricted subsets of these for which the global maximum is known.  For example, in Walsh polynomials one can select an arbitrary point and then adjust the signs of the coefficients to force this to be the global maximum \cite{tanese1989distributed}. In {\em NM} landscapes both the location and value of the global maximum is trivially known.

\subsection{Value of knowing the global minimum}
While fitnessess can be partially normalized to values $\le 1$ with Eq. \ref{NORMBYMAX} (as in Fig. \ref{searchBest}), this can still be misleading, since the range of fitness values has not been properly accounted for. It is thus preferable to normalize to values in the closed interval $[0,1]$ with Eq. (\ref{NORMBYMINMAX}), as in Fig. \ref{searchBestM1}. For example, in Fig. \ref{errorbarMinMax} we illustrate how both increase in relative fitnesses over time and the variability of fitnesses on different landscapes with the same maximum order $M$ are over-estimated when normalizing by Eq. (\ref{NORMBYMAX}), which only requires that the maximum possible fitness value be known, relative to when the data is normalized by Eq. (\ref{NORMBYMINMAX}), which requires that both the maximum and minimum possible fitness values be known.

Finding the global minimum is NP complete in {\em NK} landscapes and Walsh polynomials.  However, in certain subsets of {\em NM} landscapes (e.g., Type II and Type III {\em NM} landscapes) the value and location of the global minimum is trivially known, enabling proper normalization of fitnesses.

\subsection{Value of arbitrary arity of the alphabet}
Both {\em NK} landscapes and Walsh polynomials are defined for combinatorial problems with binary alphabets  \cite{kauffman1989nk}\cite{mitchel91}\cite{kallel2001properties}. There are also a variety of benchmark problems with tunable difficulty for real-valued alphabets (e.g., \cite{caamano2010real,ronkkonen2011framework}). However, in some applications it would be desirable to have one type of model with tunable ruggedness that could be applied to binary alphabets, integer alphabets, real-valued alphabets, or heterogeneous alphabets. For example, in real clinical fitness landscapes, decision variables can have a variety of arities ranging from binary (e.g., whether or not a certain practice is performed) to  real-valued (e.g., the amount or duration of application of a particular treatment) \cite{manukyanteam}. 

{\em NM} landscapes are well-defined for alphabets of all arities (including mixed arities); changing the arity does not change the location or value of the global maximum or minimum. 

\subsection{Value of transparency of interactions}
Various researchers are working on developing algorithms to try to detect which interactions are present in fitness landscapes and use these inferred interactions to guide the search (e.g., the linkage tree genetic algorithm \cite{thierens2013hierarchical}). Being able to easily control exactly which feature interactions are present and also know the relative strengths of these interactions would facilitate the testing and validation of such approaches, as one could easily see whether the algorithm was properly estimating interaction terms.   

{\em NK} landscapes offer little control over which interactions are to be included, and once generated it is non-obvious which interaction terms are present or what their coefficients values are (without significant effort \cite{buzas2013analysis}). Walsh polynomials present a framework where specific interaction terms can be included or excluded from the model, but the notation can be confusing and obfuscates which terms are present (e.g., see the example in Eq. (\ref{equivalence})). In {\em NM} landscapes, the interaction terms and their coefficients are obvious, since this is how interaction models are defined (e.g., see the example in Eq. (\ref{interactionExample})).

\section{Conclusions}
We propose a new class of fitness landscapes with tunable degrees of epistasis, referred to as {\em NM} landscapes. All {\em NM} landscapes have a known global optimum, various subsets of {\em NM} landscapes have a known global minimum (thus permitting proper normalization of fitness values), the ruggedness and search difficulty of {\em NM} landscapes can be made to be relatively smoothly tunable, {\em NM} landscapes are well-defined on alphabets of any arity, and which epistatic interactions are included in a particular instantiation of an {\em NM} landscape is easily controlled or analyzed. In summary, {\em NM} landscapes are a simple but powerful class of models that offer several benefits over {\em NK} landscapes and Walsh polynomials as benchmark models with tunable epistasis.


%

%

\section*{Acknowledgment}
This work was funded in part by the NIH Eunice Kennedy Shriver National Institute of Child Health \& Human Development award 1R21HD068296.

\ifCLASSOPTIONcaptionsoff
  \newpage
\fi



%

\bibliographystyle{IEEEtran}
\bibliography{TEVC-00275-2014} 

%

\begin{IEEEbiography}[{\includegraphics[width=1in]{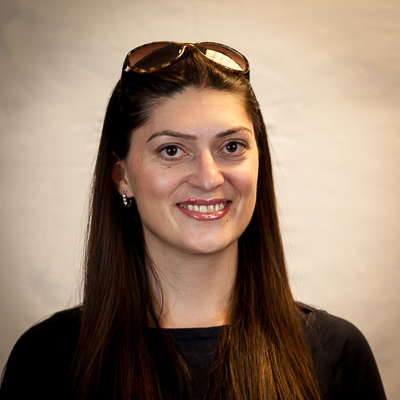}}]{Narine Manukyan}
received a BS in Applied Mathematics and Informatics at Yerevan State University in Armenia in 2008, and her MS in Computer Science at the University of Vermont in 2011. She is currently working on her PhD in Computer Science at the University of Vermont, where she also serves as treasurer of the Graduate Student Senate and Vice President of Education in the Project Management Institute Champlain chapter.  Her research interests include artificial intelligence, evolutionary computation, complex networks, data mining, machine learning and modeling complex systems.
\end{IEEEbiography}
\begin{IEEEbiography}[{\includegraphics[width=1in]{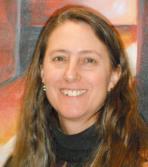}}]{Margaret J. Eppstein}
is Associate Professor and Chair of the Department of Computer Science at the University of Vermont, Burlington, VT, USA. She received the B.S. degree in zoology from Michigan State University in 1978, and the M.S. degree in computer science and the Ph.D. degree in environmental engineering from the University of Vermont, Burlington, in 1983 and 1997, respectively. She has been on the Computer Science Faculty at the University of Vermont since 1983 (Lecturer from 1983-2001; Research Assistant Professor from 1997-2002; Assistant Professor from 2002-2008; Associate Professor since 2008) and was the founding Director of the Vermont Complex Systems Center from 2006-2010. Dr. Eppstein�s current research interests involve complex systems analysis and modeling in a wide variety of application domains, including biological, environmental, technological, and social systems. 
\end{IEEEbiography}
\begin{IEEEbiography}[{\includegraphics[width=1in]{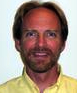}}]{Jeffrey S. Buzas}
 is Professor of Mathematics and Statistics and Director of the
Statistics Program at the University of Vermont.
Dr. Buzas received his Ph.D. in statistics at North Carolina State
University in 1993.  His recent interest in evolutionary computation
was the result of collaborations with computer scientists and physicians on
how to model the search for better medical practice and treatments
as a search on fitness landscapes.
\end{IEEEbiography}




\end{document}